\documentclass[10pt,dvipsnames,sort]{customstyle}

\usepackage{makecell} 
\usepackage{helvet}

\usepackage[
  backend=biber,
  style=numeric-comp,
  url=false,
  giveninits=true,
  uniquename=init,
  firstinits=true,
  uniquelist=false,
  maxbibnames=99,
  sorting=none
]{biblatex}
\usepackage{amssymb}
\DeclareNameAlias{sortname}{family-given}
\DeclareFieldFormat{urldate}{}
\DeclareNameAlias{default}{family-given}
\DeclareFieldFormat
[article,incollection,inproceedings,patent,thesis,unpublished]
  {title}{#1}

\renewbibmacro{in:}{}

\AtEveryBibitem{\clearfield{month}}

\AtEveryBibitem{%
  \clearfield{issn}%
  \clearfield{isbn}%
  \clearfield{note}%
  \clearfield{language}%
}

\DeclareFieldFormat{pages}{#1}
\DeclareFieldFormat{postnote}{#1}

\usepackage{makecell}
\usepackage{helvet}
\usepackage{multirow} 
\usepackage{tabularx}
\usepackage{graphicx}
\usepackage{booktabs}
\usepackage{caption}
\usepackage{wrapfig}
\usepackage{hyperref}
\usepackage{placeins}
\usepackage[utf8]{inputenc}
\usepackage{textcomp}
\usepackage{makecell}
\usepackage{booktabs}
\usepackage{longtable}
\usepackage{underscore}

\title{EDAPT: Towards Calibration-Free BCIs with Continual Online Adaptation}

\corr{lisa.haxel@uni-tuebingen.de}{}
\corr{jaivardhan.kapoor@uni-tuebingen.de}{}
\corr{jakob.macke@uni-tuebingen.de}{}

\author[1,2,3,4]{Lisa Haxel\textsuperscript{*}}
\author[1,2]{Jaivardhan Kapoor\textsuperscript{*}}
\author[3,4]{Ulf Ziemann}
\author[1,2,5,6]{Jakob H. Macke}

\affil[1]{Excellence Cluster Machine Learning, University of Tübingen, Tübingen, Germany}
\affil[2]{Tübingen AI Center, University of Tübingen, Tübingen, Germany}
\affil[3]{Department of Neurology and Stroke, University of Tübingen, Tübingen, Germany}
\affil[4]{Hertie Institute for Clinical Brain Research, University of Tübingen, Tübingen, Germany}
\affil[5]{Department of Empirical Inference, Max Planck Institute for Intelligent Systems, Tübingen, Germany}
\affil[6]{Hertie Institute for AI in Brain Health, Tübingen, Germany}

\begin{document}

\maketitle

\begin{abstract}
Brain-computer interfaces (BCIs) suffer from accuracy degradation as neural signals drift over time and vary across users, requiring frequent recalibration that limits practical deployment. We introduce EDAPT, a task- and model-agnostic framework that eliminates calibration through continual model adaptation. EDAPT first trains a baseline decoder using data from multiple users, then continually personalizes this model via supervised finetuning as the neural patterns evolve during use. We tested EDAPT across nine datasets covering three BCI tasks, and found that it consistently improved accuracy over conventional, static methods. These improvements primarily stem from combining population-level pretraining and online continual finetuning, with unsupervised domain adaptation providing further gains on some datasets. EDAPT runs efficiently, updating models within 200 milliseconds on consumer-grade hardware. Finally, decoding accuracy scales with total data budget rather than its allocation between subjects and trials. 
EDAPT provides a practical pathway toward calibration-free BCIs, reducing a major barrier to BCI deployment.
\end{abstract}

\section{Introduction}
Neurological conditions like stroke, spinal cord injury, and amyotrophic lateral sclerosis (ALS) leave millions worldwide with severe communication and mobility limitations despite intact cognitive functions \cite{WHO2019}, presenting significant challenges that Brain-computer interfaces (BCIs) are beginning to address. BCIs offer promising pathways to restore these abilities by translating brain signals into commands for assistive devices \cite{Daly2008}. Among BCI modalities, non-invasive electroencephalography (EEG) is widely used due to its portability and relatively low cost, making it a key technology for practical applications. 

Decoding user intentions from EEG signals is an active area of research. Both traditional statistical methods and deep learning have shown promise across various paradigms and tasks \cite{Varbu2022, Khademi2023, Schirrmeister2017, Lawhern2018}. Compared to traditional approaches, neural networks offer greater flexibility by minimizing feature engineering. They are more capable of extracting complex intentions from raw EEG signals, and their performance scales up when trained on more data \cite{sliwowski2023impact, levy2025braintotext}.

However, the strong offline performance of deep learning-based EEG decoders often fails to translate to real-world deployment. This gap is primarily due to the non-stationary nature of EEG signals. Neural patterns vary significantly between users and drift over time due to factors like fatigue, cognitive adaptation, or changes in electrode impedance \cite{Krumpe2017, Raza2019}. Consequently, models trained on static datasets cannot generalize to these distribution shifts, leading to performance degradation during operation \cite{Wu2017, Raza2019, Carrara2023, Wimpff2024}. To compensate, these systems require lengthy, subject-specific calibration before each session, hindering the practical adoption of BCI technology \cite{Raza2019}.

Current strategies to address this issue typically follow a two-stage process: First, multi-subject pretraining is used to create a robust initial model that generalizes well to new users \cite{levy2025braintotext, Wimpff2024Calib}. Second, from this pretrained model, online adaptation is applied to handle the signal drifts of a specific user. Real-time adaptation methods are often unsupervised; techniques like covariance alignment can mitigate distribution shifts without new labels \cite{Junqueira2024, He2020, Ouahidi2024, Wimpff2024} but, in doing so, fail to make use of valuable ground-truth labels available in many BCI settings. While supervised learning methods can leverage these labels to refine the model, they are generally applied offline---for instance, as a separate, pre-session finetuning step—rather than as a continual, real-time process \cite{Rajpura2022, Wimpff2024}. A framework that unifies these offline and online components into a single, seamless system remains an open challenge.

To address this gap, we introduce EDAPT, a task- and model-agnostic framework that transforms BCI decoding from a static prediction task into a continual online learning process. EDAPT operates in two distinct stages. First, an offline population-level pretraining step establishes a robust, general model from existing population data. Then, during online deployment, EDAPT adapts after each trial using supervised continual finetuning (CFT) on a sliding window of recent labeled data to personalize the model. As a complementary and optional step, the framework can simultaneously incorporate unsupervised domain adaptation (UDA) techniques like covariance alignment \cite{Junqueira2024} and adaptive batch normalization \cite{wimpff2025finetuning} to further counteract signal drift. This combined approach allows the decoder’s performance to improve continuously from the very first trial, eliminating the need for a separate calibration phase.

We validate EDAPT across a diverse set of nine datasets, three major BCI paradigms, and four deep learning models. Our contributions are fourfold: First, we show that our integrated approach significantly improves decoding accuracy over static models for nearly all subjects. Second, through ablation studies, we identify the combination of population-level pretraining and CFT as the primary driver of this performance gain and confirm that UDA techniques offer a complementary benefit for some datasets. Third, we validate the framework's feasibility for real-time deployment, demonstrating low latencies on consumer hardware. Finally, our scaling analysis confirms that while model performance scales with both the number of pretraining subjects and trials per subject, the key determinant of downstream performance is the total data budget, with the model being largely robust to how that budget is allocated between subjects and trials. This principle of data efficiency also applies to online adaptation, where we show that CFT matches the zero-shot baseline performance with less data requirements. Together, these results establish EDAPT as an effective and practical framework for adaptive, calibration-free BCI decoding.

\section{Results}
We conducted a comprehensive evaluation of our proposed EDAPT framework to assess its effectiveness for online, calibration-free BCI decoding. The experiments spanned three major BCI paradigms (Motor Imagery [MI], P300, and SSVEP), nine distinct datasets (three per paradigm), and four deep learning architectures to ensure the generalizability of our findings. For clarity and conciseness, the main text presents results from one representative dataset per paradigm. The complete, exhaustive results for all nine datasets are provided in the Appendix.

\subsection{EDAPT enables calibration-free BCI decoding through continual finetuning}
\begin{figure*}[t]
\centering
\includegraphics[width=\textwidth]{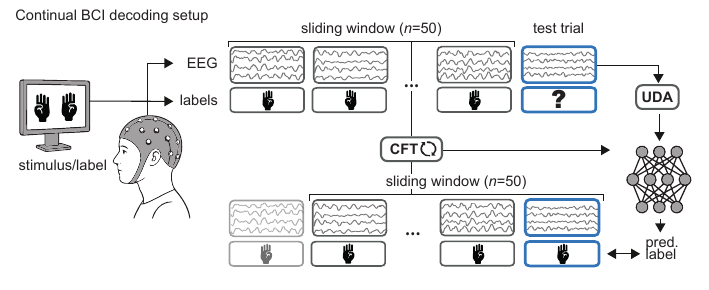}
\caption{\textbf{Overview of the EDAPT continual learning framework.} A neural network predicts the task label based on incoming EEG data from a test trial, which has optionally undergone unsupervised domain adaptation (UDA) to handle distribution shifts. After the true label is revealed, the model is updated through gradient-based continual finetuning (CFT) using a sliding window of recent EEG-label pairs. This closed-loop process allows the model to adapt to evolving neural activity patterns without a separate offline calibration phase. While illustrated using motor imagery (MI) here, EDAPT also applies to other paradigms such as P300 and SSVEP.
}
\label{fig1}
\end{figure*}

The EDAPT framework integrates its core components into a continuous, closed-loop process suitable for real-time deployment (Fig. \ref{fig1}). For each incoming trial, the model makes a prediction. Then, upon receiving the true label, it updates its parameters via continual finetuning (CFT) on a sliding window of recent, labeled trials. Optionally, before making a prediction, the model can first perform UDA to align the input data distribution. This design enables immediate use after pretraining and continuous personalization to the user without a separate calibration phase.

\FloatBarrier

\subsection{Continual finetuning consistently improves performance over static models}
\begin{figure*}[t]
\centering
\includegraphics[width=\textwidth]{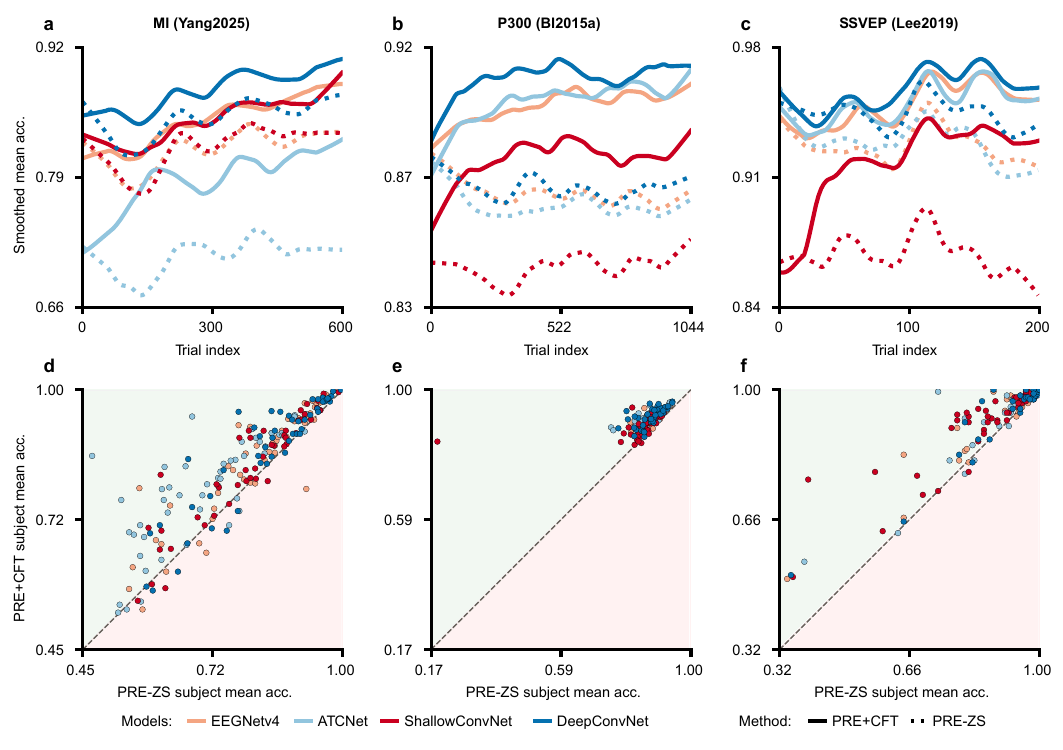}
\caption{\textbf{Continual finetuning (CFT) improves decoding accuracy over zero-shot (ZS) population-level pretraining.} We compare model performance on three BCI paradigms (columns), with different colors representing distinct model architectures.
\textbf{(a, b, c)} Smoothed trial-by-trial accuracy shows continually finetuned models (solid lines) consistently improving and outperforming static ZS baselines (dotted lines) throughout a session.
\textbf{(d, e, f)} Per-subject mean accuracy comparison. Each point represents a subject's performance with CFT (y-axis) versus ZS baseline (x-axis). Points above the identity line (green-shaded region) indicate benefit from continual finetuning.}
\label{fig2}
\end{figure*}

To evaluate the benefits of online personalization, we first compared a static, pretrained model (zero-shot, PRE-ZS) against one that continually finetunes with new data (PRE+CFT). This adaptive approach yielded substantial performance gains across all paradigms and model architectures. Trial-by-trial learning dynamics (Fig.~\ref{fig2}a-c) show that continually finetuned models (solid lines) consistently and rapidly improve in accuracy, surpassing the static PRE-ZS baselines (dotted lines).

This improvement is not just an aggregate effect but holds true on a per-subject basis. Across all three paradigms, online personalization benefits nearly every individual subject, and leads to mean decoding accuracy that is consistently higher with finetuning (y-axis) than with the static baseline (x-axis) (Fig.~\ref{fig2}d-f, green region). These findings generalize across all nine datasets we evaluated (Appendix, Fig.~\ref{figs1a}-\ref{figs1c}, column 2). 

Generally, we observed that CFT has biggest benefits on tasks that have large inter-subject variability or require continual adaptation to subject-specific patterns over a session. For instance, we observe a large increase in performance between PRE-ZS and PRE+CFT for datasets in the P300 paradigm (Fig.~\ref{fig2}b,e, Fig.~\ref{figs1b}). Decoding in P300 requires pinpointing short-lived temporal patterns with highly subject-specific signal timings (varying between 250-600ms), but are otherwise consistent within a subject. CFT allows the model be tuned to the exact timing of a test subject, significantly boosting its accuracy. Similarly, in the MI task, individuals vary in their ability for imagery and strategy, along with anatomical variability of the motor cortex. CFT adapts the pretrained model to the test subject, while improving performance steadily within a session by accounting for drift in imagery signal (Fig.~\ref{fig2}a,d, Fig.~\ref{figs1a}). In tasks where the signal-to-noise ratio is already high with little inter-subject variations (SSVEP, see Fig.~\ref{fig2}c,f, Fig.~ \ref{figs1c}), improvements are less pronounced.

\FloatBarrier

To dissect which components of the approach drive EDAPT's  performance gain, we conducted a component ablation study (Table \ref{tab:merged_table}). We considered five key configurations: the static pretrained baseline (PRE-ZS), the continually finetuned model (PRE+CFT), a model continually finetuned from random initialization for each test subject(CFT-only), a pretrained model using only unsupervised domain adaptation (PRE+UDA), and combining all three components (PRE+UDA+CFT). This design allows for a precise quantification of the performance gained at each adaptation stage. See Appendix for ablation studies exploring alternative configurations (Table \ref{tab:full_adaptation_comparison}).

In our ablation studies, we identified three key effects: First, we found that population-level pretraining is foundational for high performance. We observed that models initialized with PRE-ZS established a strong performance baseline, substantially outperforming models trained from randomly initialized models on a per-subject basis. This effect was particularly high for the SSVEP dataset, where DeepConvNet's accuracy, for instance, dropped from 0.95 to 0.32 without a pretrained initialization.

Second, from this pretrained baseline, we identified supervised continual finetuning (PRE+CFT) as the most powerful and reliable driver of performance. This single addition consistently and significantly improved decoding accuracy over the static PRE-ZS model. In contrast, we found that applying only unsupervised domain adaptation (PRE+UDA) yielded inconsistent effects; while it provided a boost for some model-dataset pairs (e.g., ATCNet on Yang2025), it offered only marginal gains or even degraded performance in others (e.g., EEGNetv4 on BI2015a, dropping from 0.87 to 0.83).

Finally, we observed paradigm-dependent synergies when combining all components into the full EDAPT model (PRE+UDA+CFT). For the event-related P300 and SSVEP paradigms, we found that the full model consistently yielded the highest or joint-highest performance, suggesting UDA is a valuable complement to CFT for these tasks. For the self-paced motor imagery (MI) paradigm, however, we noted that the simpler PRE+CFT model was often superior. We attribute this result to the possibility that for MI, the statistical alignment from UDA may distort subtle, task-relevant spatial features that are better captured by end-to-end gradient-based optimization alone.

\begin{table}[htbp]
\centering
\caption{\textbf{Population-level pretraining and continual finetuning are key drivers of BCI decoding performance.} Mean classification accuracy ($\pm$ standard deviation) for five model configurations across three paradigms. This comparison quantifies the impact of different components---population-level pretraining (PRE), supervised continual finetuning (CFT), and  unsupervised domain adaptation (UDA). For each model and dataset, the best-performing configuration is in \textbf{bold} and the second-best is \underline{underlined}. Asterisks indicate statistical significance from a one-sided paired t-test against the PRE-ZS baseline after Benjamini-Hochberg correction ($^{*}$p<0.05, $^{**}$p<0.01, $^{***}$p<0.001).}
\centering
\resizebox{\textwidth}{!}{
\begin{tabular}{lllllll} 
\toprule
Dataset & Model & PRE-ZS & CFT-only & PRE+UDA & PRE+CFT & PRE+UDA+CFT \\ 
\midrule
\multirow{4}{*}{MI (Yang2025)} & EEGNetv4 & 0.81 $\pm$ 0.12 & 0.78 $\pm$ 0.12 & 0.80 $\pm$ 0.12 & \textbf{0.85 $\pm$ 0.12$^{***}$} & \underline{0.84 $\pm$ 0.12$^{***}$} \\ 
 & ATCNet & 0.71 $\pm$ 0.12 & 0.59 $\pm$ 0.09 & 0.75 $\pm$ 0.10$^{***}$ & \textbf{0.79 $\pm$ 0.12$^{***}$} & \underline{0.78 $\pm$ 0.11$^{***}$} \\ 
 & ShallowConvNet & \underline{0.82 $\pm$ 0.12} & 0.73 $\pm$ 0.12 & 0.78 $\pm$ 0.11 & \textbf{0.85 $\pm$ 0.11$^{***}$} & \underline{0.82 $\pm$ 0.12} \\ 
 & DeepConvNet & 0.85 $\pm$ 0.12 & 0.74 $\pm$ 0.12 & 0.83 $\pm$ 0.12 & \textbf{0.88 $\pm$ 0.11$^{***}$} & \underline{0.86 $\pm$ 0.12$^{*}$} \\ 
\midrule
\multirow{4}{*}{P300 (BI2015a)} & EEGNetv4 & 0.87 $\pm$ 0.03 & 0.86 $\pm$ 0.02 & 0.83 $\pm$ 0.03 & \underline{0.90 $\pm$ 0.03$^{***}$} & \textbf{0.91 $\pm$ 0.03$^{***}$} \\ 
 & ATCNet & \underline{0.87 $\pm$ 0.04} & 0.84 $\pm$ 0.03 & 0.84 $\pm$ 0.03 & \textbf{0.91 $\pm$ 0.03$^{***}$} & \textbf{0.91 $\pm$ 0.03$^{***}$} \\ 
 & ShallowConvNet & 0.84 $\pm$ 0.11 & 0.80 $\pm$ 0.02 & 0.78 $\pm$ 0.07 & \underline{0.88 $\pm$ 0.03$^{**}$} & \textbf{0.90 $\pm$ 0.03$^{**}$} \\ 
 & DeepConvNet & \underline{0.87 $\pm$ 0.04} & 0.82 $\pm$ 0.05 & 0.83 $\pm$ 0.03 & \textbf{0.92 $\pm$ 0.03$^{***}$} & \textbf{0.92 $\pm$ 0.03$^{***}$} \\ 
\midrule
\multirow{4}{*}{SSVEP (Lee2019)} & EEGNetv4 & \underline{0.93 $\pm$ 0.12} & 0.66 $\pm$ 0.15 & \underline{0.93 $\pm$ 0.12} & \textbf{0.95 $\pm$ 0.09$^{***}$} & \textbf{0.95 $\pm$ 0.09$^{***}$} \\ 
 & ATCNet & 0.93 $\pm$ 0.11 & 0.49 $\pm$ 0.15 & 0.94 $\pm$ 0.10 & \underline{0.95 $\pm$ 0.09$^{**}$} & \textbf{0.96 $\pm$ 0.08$^{**}$} \\ 
 & ShallowConvNet & 0.87 $\pm$ 0.14 & 0.38 $\pm$ 0.12 & \underline{0.89 $\pm$ 0.14$^{*}$} & \textbf{0.92 $\pm$ 0.10$^{***}$} & \textbf{0.92 $\pm$ 0.10$^{***}$} \\ 
 & DeepConvNet & \underline{0.95 $\pm$ 0.11} & 0.32 $\pm$ 0.09 & \underline{0.95 $\pm$ 0.11} & \textbf{0.96 $\pm$ 0.09$^{**}$} & \textbf{0.96 $\pm$ 0.08$^{**}$} \\ 
\bottomrule
\end{tabular}%
}
\label{tab:merged_table}
\end{table}

\subsection{Low-latency components enable real-time BCI operation}
\begin{wrapfigure}{l}{0.5\textwidth}
\centering
\includegraphics[width=0.5\textwidth]{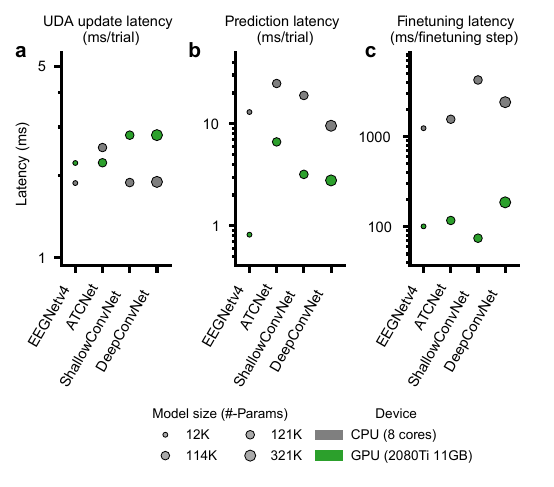}
\caption{\textbf{EDAPT components exhibit low latency suitable for real-time BCI.} Latency measurements for four deep learning architectures on the Yang2025 motor imagery dataset, benchmarked on an 8-core CPU and an NVIDIA 2080Ti (11 GB) GPU. In a typical BCI trial, a real-time sequence is executed to provide immediate feedback to the user: first the unsupervised domain adaptation (UDA) update is performed \textbf{(a)}, followed by the prediction \textbf{(b)}. The computationally most intensive continual finetuning (CFT) step \textbf{(c)} occurs during the subsequent inter-trial interval, outside the critical feedback loop. The results show that the total latency for the real-time path (the sum of (a) and (b)), especially on a consumer GPU, is well within the constraints for interactive BCI applications. The color of the circular data points indicates the device, their size is proportional to the model's parameter count, and all panels use a logarithmic y-axis.}
\label{fig3}
\end{wrapfigure}

For a continual learning framework to be practical in real-world BCI applications, its online components must operate with minimal latency. We therefore quantified the computational cost of EDAPT's three online stages—the optional unsupervised UDA statistics update, the forward pass for prediction, and a complete single-batch update for continual finetuning (CFT)—across all nine datasets. We measured wall-clock latencies on both a multi-core CPU and a consumer-grade GPU to assess real-time feasibility (Fig.~\ref{fig3}, Table~\ref{tab:latency_summary_cpu_gpu}).

The components essential for immediate user feedback, namely UDA updates and subsequent prediction, introduce negligible delay. Across all nine datasets, the GPU-accelerated prediction latency was typically under 10 ms, with the UDA statistics update generally adding less than 5ms (Fig.~\ref{fig3}, Table \ref{tab:latency_summary_cpu_gpu}). This confirms that the critical real-time path of the framework is highly responsive and suitable for interactive applications.

While the CFT update is the most computationally intensive stage, it remains feasible for online personalization when using hardware acceleration. CPU-based finetuning could take several seconds for datasets with high sampling rates, making it risky for seamless online use due to low latency margins. However, a consumer-grade GPU dramatically reduces this time, with most CFT updates completing in under 200 ms across nearly all models and datasets (Fig.~\ref{fig3}, Table \ref{tab:latency_summary_cpu_gpu}). This latency is well within the typical inter-trial interval of a BCI task, allowing for the model to be personalized asynchronously without disrupting the user experience.

\noindent
\FloatBarrier

\subsection{Performance in EDAPT scales consistently with more data}
\begin{figure*}[h!]
\centering
\includegraphics[width=\textwidth]{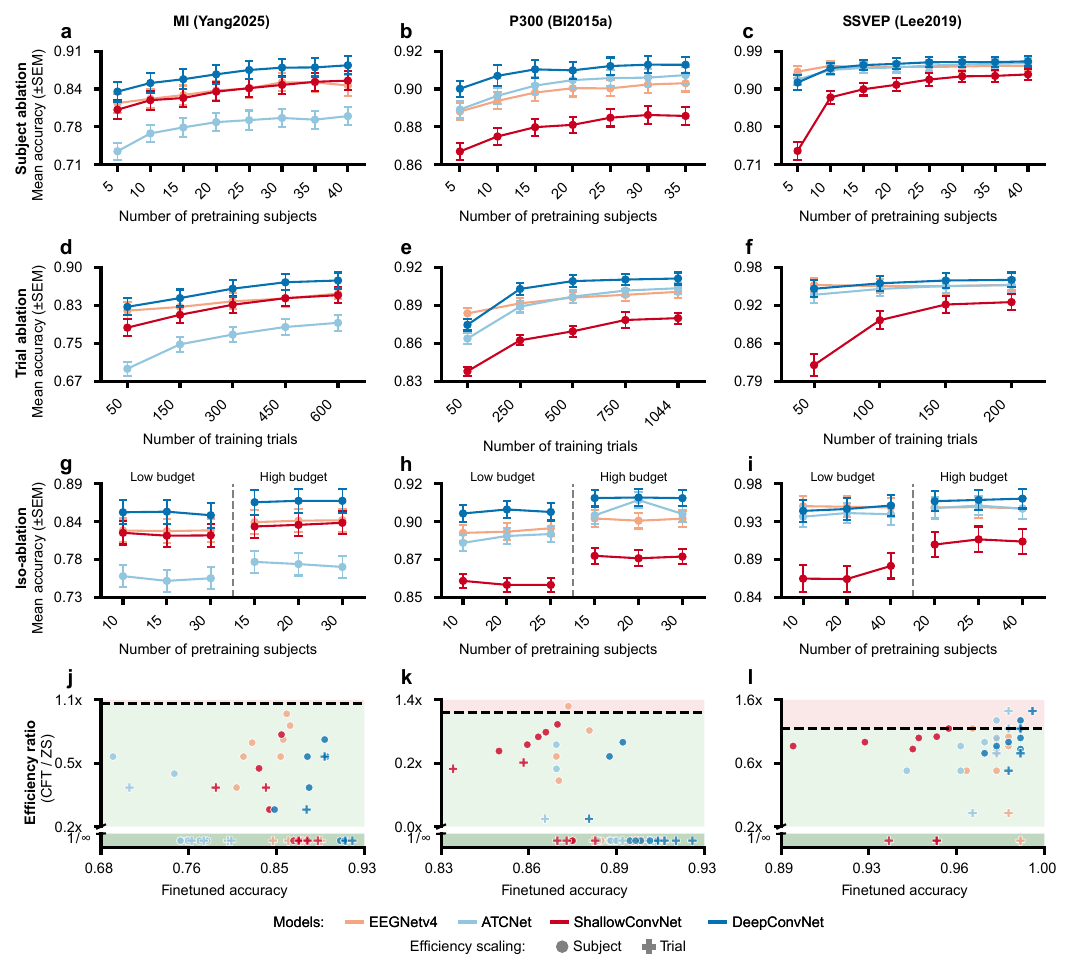}
\caption{\textbf{Performance scales with pretraining data quantity.} Scaling analysis of finetuned model accuracy across three BCI paradigms (columns) and four evaluation strategies (rows). All results show mean accuracy (±SEM) across test subjects.
\textbf{First row (a-c): Subject ablation.} Performance improves as the number of pretraining subjects increases, when holding the number of trials per subject constant.
\textbf{Second row (d-f): Trial ablation.} Performance improves with an increasing number of pretraining trials per subject, when holding the number of subjects constant.
\textbf{ Third row (g-i): Iso-ablation.} Within fixed "Low" and "High" total data budgets (e.g., for MI, Low $\approx$4.5k trials, High $\approx$9k trials), configurations range from fewer subjects with more trials (left) to more subjects with fewer trials (right). Across this range, the resulting performance is largely consistent, indicating that the total data budget is the primary driver of accuracy, not its specific composition.
\textbf{Fourth row (j-l): CFT data efficiency gains.} Y-axis shows the data ratio: how much pretraining data PRE+CFT needs relative to PRE-ZS to achieve the same accuracy (x-axis, within 1\% tolerance). Values below 1.0 (green region) mean CFT is more data-efficient. The label "0.5x" means CFT needs half the data; "1/$\infty$" means PRE-ZS cannot reach that accuracy regardless of pretraining amount.
}
\label{fig4}
\end{figure*}

We conducted a systematic scaling analysis to investigate the impact of the number of pretraining subjects, the number of trials per  pretraining subject, and the trade-off between these factors under a fixed data budget. The results for continually finetuned (CFT)- (Fig.~\ref{fig4}, Fig.~\ref{figs2a}-\ref{figs2a}) and zero-shot population-level pretraining models Fig.~\ref{figs2d}-\ref{figs2f}) demonstrate similar scaling trends.

First, we found that performance scales monotonically with the total amount of pretraining data available. Increasing either the number of subjects in the pretraining pool (Fig.~\ref{fig4}, Fig.~\ref{figs2a}-\ref{figs2f}, top row) or the per-subject longitudinal data in the form of trials per session (Fig.~\ref{fig4}, Fig.~\ref{figs2a}-\ref{figs2f}, second row) yields clear and consistent improvements in decoding accuracy across all paradigms. In all cases, the continually finetuned models outperform their zero-shot counterparts, highlighting the benefit of online personalization regardless of the initial pretraining data quantity (Fig.~\ref{figs2a}-\ref{figs2f}).

We then investigated the trade-off between distributing data between multiple subjects vs. longitudinal data per session for each subject, given a fixed total budget over trials (Fig.~\ref{fig4}, Fig.~\ref{figs2a}-\ref{figs2f}, third row). This iso-scaling analysis was performed under two budget constraints for each dataset, a "Low" and "High" total number of pretraining trials. For the representative datasets, we considered budgets of $\approx$4,500 and $\approx$9,000 total trials for Yang2025 (MI) and BI2015a (P300), and $\approx$2,000 and $\approx$4,000 trials for Lee2019 (SSVEP). Our results reveal that within a constant total number of pretraining trials, configurations using fewer subjects with more trials performed similarly to those with more subjects and fewer trials (Fig.~\ref{fig4}, Fig.~\ref{figs2a}-\ref{figs2f}). This finding suggests that the total number of trials in the pretraining pool is the primary driver of performance, not its specific composition. 

Finally, our analysis demonstrates that CFT substantially increases the data efficiency of the pretrained model. We found that to reach any given level of accuracy, the PRE+CFT configuration mostly required substantially less pretraining data—fewer subjects or fewer trials per subject—than the static zero-shot (PRE-ZS) model (Fig.~\ref{fig4},~\ref{figs2a}-\ref{figs2f}, bottom row, green region with p<0.01). Furthermore, in many cases, online adaptation boosted performance to levels that the static model could not reach, even when using the entire pretraining dataset. Therefore, applying CFT not only personalizes a model but also serves as a powerful method to maximize the value of a limited pretraining dataset for EEG decoding.

\FloatBarrier

\section{Discussion}
We introduced EDAPT, a modular framework designed to address the critical challenge of calibration in brain-computer interfaces. By combining population-level pretraining with online continual finetuning (CFT) and optional unsupervised domain adaptation (UDA), we demonstrated a practical path toward calibration-free BCI decoding. Our experiments, spanning nine datasets and three major paradigms, provide three key takeaways. First, combining a strong pretrained initialization and trial-wise CFT consistently and substantially improves performance over static models, effectively personalizing the decoding model to a new user in real-time. Second, EDAPT is computationally efficient, with wall-clock latencies suitable for consumer hardware, hence ensuring that online learning does not disrupt the user experience. Third, our scaling analysis offers two practical insights for BCI adaptation---for a fixed data budget,
model performance primarily driven by the total pretraining data budget, not its specific composition (i.e., the trade-off between more subjects or more trials),  which allows researchers to design flexible data collection strategies based on their specific constraints. In addition, trial-wise CFT significantly reduces the amount of pretraining data required to achieve a target performance. Together, these findings establish that continual online learning is not only feasible, but is a highly effective strategy for creating high-performance, user-adaptive BCIs.

Our work establishes a strong foundation for online adaptation, but its limitations highlight important pathways for future research. The most significant limitation is the transition from a simulated open-loop setting to a true closed-loop BCI system. Factors like user-in-the-loop co-adaptation, shifts in control strategy, or changes in physiological state are not just weaknesses of our method, but fundamental challenges for the entire BCI field \cite{Murphy2016}. Indeed, EDAPT is designed precisely to contend with such real-world changes. Therefore, a clear next step is to conduct closed-loop live studies to quantify how EDAPT's accuracy gains translate into improvements in user performance, cognitive workload, and overall satisfaction.

Additionally, our framework's reliance on supervised signals for CFT limits its current applicability to cue-based paradigms where ground-truth labels are readily available. Extending to more autonomous, self-paced scenarios would require addressing the fundamental challenge of label scarcity in online BCI settings.

The core strength of the EDAPT framework lies in its modular design that enables flexible combination of components while maintaining robust performance through their synergistic interaction. Its design, for instance, inherently mitigates a key risk of transfer learning, as CFT rapidly corrects for initial errors from a potentially mismatched pretraining dataset. Furthermore, its modularity enables integration with existing UDA techniques—even though the UDA components we tested showed inconsistent benefits, EDAPT can readily incorporate more sophisticated unsupervised methods as they mature. This positions EDAPT as a general blueprint for a new class of adaptive BCIs. The practical impact of EDAPT varies across performance regimes. While absolute accuracy improvements are naturally smaller at high baseline performance levels (consistent with our results), even modest gains become increasingly valuable as systems approach clinical deployment thresholds. For instance, improving a BCI's accuracy from 90\% to 95\% cuts the system's error rate in half (from 10\% to 5\%), though such improvements require increasingly sophisticated methods. By providing a systematic solution to the calibration bottleneck across diverse performance levels, EDAPT helps transition neurotechnology from experimental research toward more robust clinical tools.

\section{Acknowledgments}
This work was supported by the Else Kröner Medical Scientists Kolleg Clinbrain: Artificial Intelligence for Clinical Brain Research,  the Machine Learning Cluster of Excellence, EXC number 2064/1–390727645, the German Federal Ministry of Education and Research (BMBF): Tübingen AI Center, FKZ: 01IS18039A and the European Research Council (ERC Synergy) under the European Union’s Horizon 2020 research and innovation program (ConnectToBrain, grant number 810377).

\section{Author contributions}

Conceptualization, Methodology: L.H., J.K., J.H.M.
Software and Investigation: L.H., J.K.
Analysis: L.H., J.K.
Writing: L.H., J.K.
Writing (Review \& Editing): U.Z., J.H.M.
Funding acquisition: U.Z., J.H.M.
Supervision: U.Z., J.H.M.

\printbibliography
\pagebreak

\section{Methods}

\subsection{Code availability}
All code to generate results and figures is available at \href{https://github.com/mackelab/EDAPT}{\texttt{https://github.com/mackelab/EDAPT}}.

\subsection{Datasets}
We evaluated our framework on nine public EEG datasets spanning three distinct brain-computer interface (BCI) paradigms (Table~\ref{tab:datasets_final}). With the exception of Yang2025, all data were accessed using the MOABB framework \cite{Aristimunha2023}.

\paragraph{Motor Imagery (MI)}
This paradigm involves the mental rehearsal of a motor action without any overt muscle activation and is fundamental in neuroscience for investigating motor control and learning  \cite{Singh2021}. We used three MI datasets: Yang2025 \cite{Yang2025}, Lee2019\_MI \cite{Lee2019}, and BNCI2014\_001 \cite{Tangermann2012}.

\paragraph{P300 Event-Related Potential (ERP)}
This paradigm focuses on detecting a target stimulus by identifying the P3b neural response, which is a positive voltage peak occurring approximately 300\,ms after a rare, task-relevant event \cite{Kleih2011}. The task is a binary classification of target versus non-target trials. We used three P300 datasets: BI2015a \cite{Korczowski2019}, Huebner2017 \cite{Hubner2017}, and Huebner2018 \cite{Huebner2018}.

\paragraph{Steady-State Visually Evoked Potential (SSVEP)}
This paradigm requires classifying the frequency of a flickering visual stimulus at which a user is gazing. The classification is based on identifying the corresponding frequency component and its harmonics in occipital EEG channels \cite{Liu2022}. We included three SSVEP datasets: Lee2019\_SSVEP \cite{Lee2019}, MAMEM2 \cite{Oikonomou2016}, and Kalunga2016 \cite{Kalunga2016}.

\paragraph{Data preprocessing}
For each subject, data from all available sessions were concatenated to form a single recording. A uniform preprocessing pipeline was applied, consisting of a 2--47\,Hz band-pass filter and an average re-referencing of the EEG signals.

\begin{table}[h!]
\centering
\caption{Overview of the EEG datasets used in this study. Columns denote: Number of Subjects (\# Subj.), Number of Channels (\# Chan.), Sampling Rate (Fs), Number of Classes (\# Classes), Number of trials per class per session (Trials/Class/Session), Trial length in seconds (Trial len. (s)), and Number of Sessions (\# Sessions). For P300 paradigms, trial counts are given as Target (T) / Non-Target (NT) epochs per session.}
\label{tab:datasets_final}
\resizebox{\textwidth}{!}{%
\begin{tabular}{@{}llcccccccc@{}}
\toprule
\textbf{Dataset} & \textbf{Paradigm} & \textbf{\# Subj.} & \textbf{\# Chan.} & \textbf{Fs (Hz)} & \textbf{\# Classes} & \textbf{Trials/Class/Session} & \textbf{Trial len. (s)} & \textbf{\# Sessions} \\ \midrule
Yang2025 \cite{Yang2025}         & MI    & 51 & 59 & 1000 & 2 & 100                 & 4.0  & 3 \\
Lee2019\_MI \cite{Lee2019}              & MI    & 54 & 62 & 1000 & 1 & 100                 & 4.0  & 2 \\
BNCI2014\_001 \cite{Tangermann2012}  & MI    & 9  & 22 & 250  & 4 & 144                 & 4.0  & 2 \\
BI2015a \cite{Korczowski2019}                   & P300  & 43 & 32 & 512  & 2 & 825 (T) / 4131 (NT) & 1.0  & 3 \\
Huebner2017 \cite{Hubner2017}         & P300  & 13 & 31 & 1000 & 2 & 112 (T) / 364 (NT)  & 0.9  & 3 \\
Huebner2018 \cite{Huebner2018}         & P300  & 12 & 31 & 1000 & 2 & 112 (T) / 364 (NT)  & 0.9  & 3 \\
Lee2019\_SSVEP \cite{Lee2019}           & SSVEP & 54 & 62 & 1000 & 4 & 50                  & 4.0  & 1 \\
MAMEM2 \cite{Oikonomou2016}               & SSVEP & 10 & 256& 250  & 5 & 20--30              & 3.0  & 1 \\
Kalunga2016 \cite{Kalunga2016}        & SSVEP & 12 & 8  & 256  & 4 & 16                  & 2.0  & 1 \\ \bottomrule
\end{tabular}%
}
\end{table}

\subsection{Deep learning architectures}
Our evaluation included four established convolutional neural network (CNN) architectures for decoding EEG signals.
\begin{itemize}
\item \textbf{ShallowConvNet} \cite{Schirrmeister2017} consists of a temporal convolutional layer followed by a spatial filtering layer to extract features analogous to those from a filter bank common spatial pattern (FBCSP) pipeline.

\item \textbf{DeepConvNet} \cite{Schirrmeister2017} utilizes a deeper, multi-block structure composed of stacked convolutional layers. This design facilitates hierarchical feature learning by progressively extracting more abstract representations from the raw signal.

\item \textbf{EEGNet} \cite{Lawhern2018} is a compact architecture that employs depthwise separable convolutions to factorize the learning of temporal and spatial filters, thereby significantly reducing the number of trainable parameters.

\item\textbf{ATCNet} \cite{Altaheri2023} augments a convolutional feature extractor with a multi-head self-attention block and a temporal convolutional network (TCN). This hybrid structure allows the model to weigh the importance of learned features and capture long-range temporal dependencies in the EEG signal.

\end{itemize}

\subsection{Training and online adaptation framework}
Our method comprises a two-stage framework designed to first learn generalizable neural representations from a source population and then rapidly adapt these representations to a new target subject in a continuous, trial-by-trial manner.

\paragraph{Population-level pretraining}
The initial stage aims to learn a robust mapping from EEG signals to task-related classes by training on a diverse population of subjects. We employ a 2-fold cross-validation scheme, where for each fold, the model is trained on data from 50\% of the subjects and subsequently evaluated on the remaining 50\%. Before data from the training subjects is aggregated, we apply an optional preprocessing step to reduce inter-subject variability by independently aligning each subject's data using a covariance transformation matrix using that subject's trials. The model is then pretrained for 100 epochs using the Adam optimizer and a standard cross-entropy loss function (Table~\ref{tab:training_classification}).

\paragraph{Trial-by-trial online adaptation}
In the second stage, we consider a realistic, calibration-free scenario where a pretrained model is deployed on a new subject. The adaptation proceeds trial-by-trial with the following sequence: (1) optionally, for each incoming trial, unsupervised domain adaptation techniques are applied to align the input data and the model's internal features to the new subject's statistics; (2) a prediction is made for that trial; and (3) after the prediction, the model's weights are updated through supervised finetuning with the true label.

\paragraph{Unsupervised covariance alignment}
To minimize the distribution shift between the source population and the target subject, we standardize the second-order statistics of the input data \cite{He2019, Zanini2017, Kostas2020}. Each incoming trial $X \in \mathbb{R}^{C \times T}$ (where $C$ is the number of channels and $T$ is the number of time points) is transformed to an aligned trial $\tilde{X} = M \cdot X$. The transformation matrix $M = C_{\text{ref}}^{-1/2}$ is the inverse square root of a reference covariance matrix, which whitens the data with respect to this reference. Instead of recomputing a mean covariance from a buffer of past trials, we update the reference covariance $C_{\text{ref}}$ after each trial using an Exponential Moving Average (EMA):
\begin{equation}
    C_{\text{ref}}^{(t)} = \beta \cdot C_{\text{ref}}^{(t-1)} + (1-\beta) \cdot C_{\text{trial}}^{(t)}
\end{equation}
where $C_{\text{trial}}^{(t)}$ is the covariance of the current trial. The decay factor $\beta$ is set to 0.9, balancing historical and instantaneous statistics and preventing large jumps in the estimate. Our framework supports both the Euclidean and the geometrically-aware Riemannian mean in tangent space for estimating the covariance.

\paragraph{Adaptive batch normalization (AdaBN)}
Following input alignment, we adapt the model's internal feature representations using a per-trial online variant of adaptive batch normalization (AdaBN) \cite{Ioffe2015, Li2016}. Whereas standard inference relies on static population statistics (mean and variance) aggregated during pretraining, our approach instantaneously adapts the model to the target subject. For each incoming trial, the normalization statistics for every batch normalization layer are computed de novo, using only the feature activations from that single trial. This forces the model's internal feature space to be continually re-normalized with respect to the immediate statistical context of the data being processed. This  method corresponds to the `BN-1` formulation discussed in prior BCI literature \cite{Schneider2020, Wimpff2024Calib}, ensuring maximal adaptation to the immediate context of the data being processed.

\paragraph{Supervised continual finetuning (CFT)}
After the unsupervised adaptation and subsequent prediction for a given trial, the model's weights are updated using the trial's true label. This supervised learning process begins after an initial warm-up period of 20 trials, during which data is only collected and the model's weights remain frozen. For each subsequent trial, its data and label are added to a sliding window buffer that stores the 50 most recent trials. The model is then finetuned for 3 epochs using the data within this buffer. We implement two distinct finetuning strategies:
\begin{itemize}
    \item \textbf{Full finetuning:} All model parameters are updated, allowing the entire network, from feature extraction to classification, to adapt to the new subject.
    \item \textbf{Decision-only finetuning:} To reduce the risk of catastrophic forgetting and promote stability, this strategy freezes the convolutional feature extraction layers and updates only the parameters of the final, fully-connected classification layer. This focuses adaptation on the decision boundary while preserving the robust features learned during pretraining.
\end{itemize}

\begin{table}[h!]
\centering
\caption{Training configuration across the two learning phases for the classification tasks. The Adam optimizer was used with a constant learning rate. Abbreviations: LR = Learning rate, CV = Cross-Validation.}
\label{tab:training_classification}
\small %
\begin{tabularx}{\textwidth}{@{}>{\raggedright\arraybackslash}X >{\centering\arraybackslash}X >{\centering\arraybackslash}X@{}}
\toprule
\textbf{Parameter} & \textbf{Population-level pretraining} & \textbf{Online adaptation} \\
\midrule
\multicolumn{3}{l}{\textbf{Dataset settings}} \\
Subject split & 2-fold CV (50\% train / 50\% test) & test split subjects (evaluated individually) \\
trials used & all available from training subjects & sliding window of 50 most recent trials \\
\midrule
\multicolumn{3}{l}{\textbf{Training hyperparameters}} \\
Epochs & 100 & 3 per incoming trial \\
Batch size & 64 & 50 (or size of current window) \\
Learning rate (LR) & $1\times10^{-4}$ & $1\times10^{-4}$ \\
Warm-up trials & N/A & 20 \\
\midrule
\multicolumn{3}{l}{\textbf{Optimization settings}} \\
Optimizer & Adam & Adam \\
Weight decay & 0.0 & 0.0 \\
\midrule
\multicolumn{3}{l}{\textbf{Loss function}} \\
Loss & cross-entropy & cross-entropy \\
\bottomrule
\end{tabularx}
\end{table}

\subsection{Evaluation metrics}
Model performance is evaluated using two primary metrics: the zero-shot accuracy and the final overall accuracy. The zero-shot accuracy establishes a baseline by evaluating the pretrained model on all trials of a new subject without any adaptation. To analyze learning dynamics during the online phase, we track both trial-by-trial performance and a cumulative accuracy updated after each trial. The final overall accuracy, our main reported metric, is the value of this cumulative accuracy at the end of the online simulation. Final scores are obtained by averaging these per-subject metrics across test folds.

\section{Experiments}

\subsection{Component ablation study}
To systematically dissect the contributions of our framework's components, we conducted a comprehensive ablation study across all nine datasets and four deep learning architectures. The study was designed to quantify the individual and combined impact of population-level pretraining (PRE), unsupervised domain adaptation (UDA), and supervised continual finetuning (CFT). To identify the best-performing configurations for each dataset and model, the final accuracy of each setup was compared against the Zero-shot population-level pretraining (PRE-ZS) baseline. Statistical significance was determined using a one-sided paired t-test, with p-values corrected for multiple comparisons across configurations using the Benjamini-Hochberg procedure ($\ast p < 0.05$, $\ast\ast p < 0.01$, $\ast\ast\ast p < 0.001$).

\begin{itemize}
    \item{\textbf{From-scratch (CFT-only)}}
    This configuration establishes a baseline for single-subject performance by omitting population-level pretraining entirely. For each new subject, the model is initialized with random weights and trained exclusively in an online, trial-by-trial manner using our supervised continual finetuning CFT) protocol. This measures the effectiveness of online learning without the benefit of transferred knowledge.

    \item{\textbf{Zero-shot population-level pretraining (PRE-ZS)}}
    This configuration serves as our primary population-level pretraining baseline. A model is pretrained on the source population data and then evaluated on a new target subject without any subsequent online adaptation. The model's weights remain frozen. This isolates the out-of-the-box generalization performance gained from population-level pretraining.

    \item{\textbf{PRE with unsupervised adaptation (PRE+UDA)}}
    Building on the pretrained PRE-ZS model, this configuration incorporates our unsupervised domain adaptation module. For each incoming trial, it applies unsupervised covariance alignment and adaptive batch normalization (AdaBN), but no supervised weight updates are performed. This isolates the impact of purely unsupervised adaptation techniques on a pretrained model.

    \item{\textbf{EDAPT (PRE+CFT)}}
    This configuration represents our core proposed framework. It applies the continual finetuning (CFT) protocol directly to the pretrained PRE-ZS model without any unsupervised adaptation, isolating the performance gains from combining pretraining with supervised online learning.
    
    \item{\textbf{EDAPT with unsupervised adaptation (PRE+UDA+CFT)}}
    This configuration investigates the effect of adding UDA to our core framework. It begins with the pretrained model and integrates all components: the model first performs UDA on the input trial, and after a prediction is made, its weights are updated via CFT. This allows us to measure any synergistic or complementary benefits from combining both adaptation strategies.
\end{itemize}

\subsection{Additional ablations}
To further probe the mechanics of our framework, we assessed several variants of the full model.

\begin{itemize}
    \item{\textbf{Ablating pretraining (UDA+CFT)}}
    To determine if UDA is beneficial without robust pretrained features, this configuration combines both adaptation strategies but omits population-level pretraining. Model weights are randomly initialized, and for each trial, the model applies UDA before being updated with CFT.
    
    \item{\textbf{Ablating full finetuning (PRE+UDA+CFT-Dec)}}
    We explored a more constrained finetuning strategy to enhance stability. In this variant of the full framework, the supervised CFT step only updates the parameters of the final, fully-connected classification layer (decoder). The pretrained convolutional feature extraction layers remain frozen, mitigating the risk of catastrophic forgetting.
    
    \item{\textbf{Ablating adaptive batch normalization (PRE+CovAlign+CFT)}}
    To isolate the contribution of AdaBN, this ablation removes it from the full framework's UDA module. The model performs only input-level unsupervised covariance alignment before the supervised CFT step.
    
    \item{\textbf{Riemannian alignment (PRE+UDA-Riem+CFT)}}
    As a final ablation, we replaced the default Euclidean-based covariance alignment in the full framework with a geometrically-aware Riemannian alignment method \cite{Zanini2017}. This variant updates the reference covariance matrix using the Riemannian mean, which may offer a more robust handling of the underlying geometry of the data.
\end{itemize}

\subsection{Latency analysis}
To assess the real-time feasibility of our framework, we quantified the computational latency of its core online components. All benchmarks were conducted in a PyTorch environment on two distinct hardware platforms: a multi-core CPU (8-core AMD EPYC 7302) and a consumer-grade GPU (NVIDIA GeForce RTX 2080Ti with 11 GB of VRAM).

We measured the wall-clock time in milliseconds for three distinct computational stages. For GPU benchmarks, accurate timings were ensured by synchronizing the device before and after each measurement to account for the completion of any asynchronous operations. The three measured stages were:
\begin{itemize}
    \item \textbf{UDA update latency:} The time required to update the unsupervised domain adaptation statistics using a single incoming trial. This was measured over 100 repetitions, preceded by 10 warm-up runs to stabilize system state.
    \item \textbf{Prediction latency:} The time for a single, full forward pass, including the application of the UDA transformations to the input and internal features. This was also measured across 100 repetitions after 10 warm-up runs.
    \item \textbf{CFT update latency:} The time to perform one complete supervised finetuning step. This  measurement includes the forward pass, loss calculation, backpropagation, and optimizer weight update for a single batch of data. The latency was averaged over 20 consecutive update steps, following an initial warm-up of 5 batch updates. A batch size of 50 trials was used, consistent with the online adaptation protocol.
\end{itemize}

This evaluation protocol was executed for all four deep learning architectures (ShallowConvNet, EEGNetv4, ATCNet, and DeepConvNet) across all nine datasets. For each dataset, latency was measured using data from the first available subject. We report the median latency across repetitions for each configuration.

\subsection{Scaling analysis}
To understand the relationship between pretraining data composition and downstream performance, we conducted a series of scaling analyses. The investigations were performed for all four model architectures on both continually finetuned and zero-shot models. Each analysis involved systematically varying the quantity and structure of the pretraining data and evaluating the resulting model's mean accuracy (±SEM) on held-out test subjects. Three distinct ablation strategies were employed:

\begin{itemize}
    \item \textbf{Subject ablation:} To assess the impact of the number of source subjects, we varied the size of the pretraining subject pool while keeping the number of trials used per subject at the maximum available for the given dataset (Table~\ref{tab:subject_ablation_configs} in Appendix).
    \item \textbf{Trial ablation:} We varied the number of pretraining trials used per subject to evaluate the influence of pretraining data depth. Throughout this ablation, the total number of pretraining subjects and the number of finetuning trials for each test subject were held constant at the maximum available. (Table~\ref{tab:trial_ablation_configs} in Appendix).
    \item \textbf{Data budget trade-off (iso-ablation):} To investigate the trade-off between subject quantity and trial depth, we evaluated performance under fixed total data budgets. For each dataset, we defined "Low" and "High" budget conditions, each corresponding to a fixed total number of pretraining trials. Within each budget, we created configurations that traded off fewer subjects with more trials against more subjects with fewer trials (Table~\ref{tab:iso_ablation_configs} in Appendix).
\end{itemize}

For the subject and iso-ablation analyses, which could involve a large proportion of the total subjects for pretraining, we employed a 5-fold cross-validation scheme. This contrasts with the 2-fold CV used in other experiments and ensures that a sufficient number of held-out subjects are available for robust evaluation in each fold, even when the pretraining set is large.

\begin{table}[htbp]
\centering
\caption{\textbf{Complete results for the EDAPT framework ablation study.} The table presents the classification accuracy (mean ± std. dev.) for four neural network models across nine datasets from three BCI paradigms. Each model is evaluated under ten different adaptation configurations. For each model-dataset combination (row), the highest accuracy is marked in \textbf{bold}, and the second-highest is \underline{underlined}. Statistical significance is calculated using a one-sided paired t-test comparing each configuration against the PRE-ZS baseline for that row. Asterisks indicate p-values after Benjamini-Hochberg correction ($^{}$p<0.05, $^{}$p<0.01, $^{***}$p<0.001).}
\resizebox{\textwidth}{!}{
\begin{tabular}{lllllllllll} 
\toprule
Dataset & Model & PRE-ZS & CFT-only & PRE+UDA & UDA+CFT-only & PRE+CFT & PRE+UDA+CFT & \makecell{PRE+UDA\\+CFT-Dec} & \makecell{PRE+UDA-\\noAdaBN+CFT} & \makecell{PRE+UDA-\\Riem+CFT} \\ 
\midrule
\multirow{4}{*}{\makecell[l]{MI\\(Yang2025)}} & EEGNetv4 & 0.816 $\pm$ 0.121 & 0.778 $\pm$ 0.116 & 0.803 $\pm$ 0.120 & 0.744 $\pm$ 0.113 & \textbf{0.847 $\pm$ 0.119$^{***}$} & \underline{0.846 $\pm$ 0.118$^{***}$} & 0.825 $\pm$ 0.120 & 0.845 $\pm$ 0.123$^{***}$ & 0.817 $\pm$ 0.114 \\ 
 & ATCNet & 0.711 $\pm$ 0.122 & 0.584 $\pm$ 0.091 & 0.752 $\pm$ 0.101$^{***}$ & 0.516 $\pm$ 0.045 & \textbf{0.791 $\pm$ 0.114$^{***}$} & \underline{0.786 $\pm$ 0.111$^{***}$} & 0.759 $\pm$ 0.101$^{***}$ & 0.786 $\pm$ 0.114$^{***}$ & 0.716 $\pm$ 0.089 \\ 
 & ShallowConvNet & 0.819 $\pm$ 0.118 & 0.736 $\pm$ 0.117 & 0.785 $\pm$ 0.111 & 0.666 $\pm$ 0.122 & \textbf{0.848 $\pm$ 0.113$^{***}$} & 0.821 $\pm$ 0.119 & 0.805 $\pm$ 0.117 & 0.822 $\pm$ 0.117 & \underline{0.824 $\pm$ 0.118} \\ 
 & DeepConvNet & 0.847 $\pm$ 0.117 & 0.744 $\pm$ 0.119 & 0.832 $\pm$ 0.118 & 0.724 $\pm$ 0.114 & \textbf{0.878 $\pm$ 0.115$^{***}$} & 0.863 $\pm$ 0.121$^{*}$ & 0.847 $\pm$ 0.118 & \underline{0.866 $\pm$ 0.119$^{*}$} & 0.856 $\pm$ 0.119 \\ 
\midrule
\multirow{4}{*}{\makecell[l]{MI\\(Lee2019)}} & EEGNetv4 & 0.760 $\pm$ 0.093 & 0.546 $\pm$ 0.070 & 0.767 $\pm$ 0.079 & 0.564 $\pm$ 0.072 & 0.778 $\pm$ 0.096$^{**}$ & 0.792 $\pm$ 0.084$^{***}$ & 0.778 $\pm$ 0.086$^{*}$ & \textbf{0.799 $\pm$ 0.083$^{***}$} & \underline{0.795 $\pm$ 0.080$^{***}$} \\ 
 & ATCNet & 0.729 $\pm$ 0.112 & 0.522 $\pm$ 0.059 & 0.744 $\pm$ 0.115 & 0.509 $\pm$ 0.044 & 0.761 $\pm$ 0.118$^{***}$ & \underline{0.774 $\pm$ 0.112$^{***}$} & 0.756 $\pm$ 0.113$^{**}$ & \textbf{0.780 $\pm$ 0.115$^{***}$} & 0.766 $\pm$ 0.117$^{***}$ \\ 
 & ShallowConvNet & 0.722 $\pm$ 0.102 & 0.574 $\pm$ 0.073 & 0.761 $\pm$ 0.083$^{***}$ & 0.565 $\pm$ 0.075 & 0.771 $\pm$ 0.097$^{***}$ & \textbf{0.784 $\pm$ 0.089$^{***}$} & 0.774 $\pm$ 0.093$^{***}$ & \underline{0.784 $\pm$ 0.087$^{**}$} & 0.772 $\pm$ 0.083$^{***}$ \\ 
 & DeepConvNet & 0.780 $\pm$ 0.108 & 0.541 $\pm$ 0.068 & 0.820 $\pm$ 0.086$^{***}$ & 0.556 $\pm$ 0.063 & 0.822 $\pm$ 0.095$^{***}$ & \textbf{0.834 $\pm$ 0.082$^{***}$} & 0.797 $\pm$ 0.094 & \underline{0.826 $\pm$ 0.082$^{***}$} & 0.810 $\pm$ 0.092$^{**}$ \\ 
\midrule
\multirow{4}{*}{\makecell[l]{MI\\(BNCI2014\_001)}} & EEGNetv4 & 0.472 $\pm$ 0.112 & 0.439 $\pm$ 0.054 & 0.513 $\pm$ 0.106 & 0.474 $\pm$ 0.059 & 0.668 $\pm$ 0.120$^{***}$ & \textbf{0.733 $\pm$ 0.103$^{***}$} & 0.633 $\pm$ 0.112$^{***}$ & 0.724 $\pm$ 0.106$^{***}$ & \underline{0.731 $\pm$ 0.110$^{***}$} \\ 
 & ATCNet & 0.464 $\pm$ 0.161 & 0.471 $\pm$ 0.087 & 0.491 $\pm$ 0.183 & 0.513 $\pm$ 0.109 & 0.708 $\pm$ 0.109$^{***}$ & \textbf{0.768 $\pm$ 0.094$^{***}$} & 0.571 $\pm$ 0.131$^{**}$ & \underline{0.768 $\pm$ 0.093$^{***}$} & 0.761 $\pm$ 0.094$^{***}$ \\ 
 & ShallowConvNet & 0.438 $\pm$ 0.137 & 0.530 $\pm$ 0.133$^{*}$ & 0.457 $\pm$ 0.162 & 0.593 $\pm$ 0.134$^{***}$ & 0.624 $\pm$ 0.149$^{***}$ & \textbf{0.732 $\pm$ 0.112$^{***}$} & 0.599 $\pm$ 0.122$^{***}$ & \textbf{0.732 $\pm$ 0.112$^{***}$} & \underline{0.730 $\pm$ 0.108$^{***}$} \\ 
 & DeepConvNet & 0.440 $\pm$ 0.149 & 0.273 $\pm$ 0.021 & 0.473 $\pm$ 0.130 & 0.276 $\pm$ 0.026 & 0.578 $\pm$ 0.094$^{***}$ & \textbf{0.639 $\pm$ 0.103$^{***}$} & 0.465 $\pm$ 0.063 & 0.632 $\pm$ 0.106$^{***}$ & \underline{0.634 $\pm$ 0.105$^{***}$} \\ 
\midrule
\multirow{4}{*}{\makecell[l]{P300\\(BI2015a)}} & EEGNetv4 & 0.866 $\pm$ 0.032 & 0.857 $\pm$ 0.023 & 0.834 $\pm$ 0.028 & 0.877 $\pm$ 0.024$^{**}$ & 0.902 $\pm$ 0.031$^{***}$ & 0.910 $\pm$ 0.027$^{***}$ & 0.880 $\pm$ 0.027$^{***}$ & \underline{0.910 $\pm$ 0.027$^{***}$} & \textbf{0.915 $\pm$ 0.026$^{***}$} \\ 
 & ATCNet & 0.864 $\pm$ 0.041 & 0.842 $\pm$ 0.026 & 0.837 $\pm$ 0.027 & 0.848 $\pm$ 0.032 & 0.906 $\pm$ 0.032$^{***}$ & \underline{0.912 $\pm$ 0.028$^{***}$} & 0.846 $\pm$ 0.035 & \underline{0.912 $\pm$ 0.028$^{***}$} & \textbf{0.917 $\pm$ 0.029$^{***}$} \\ 
 & ShallowConvNet & 0.840 $\pm$ 0.106 & 0.803 $\pm$ 0.019 & 0.782 $\pm$ 0.070 & 0.839 $\pm$ 0.020 & 0.884 $\pm$ 0.031$^{**}$ & \underline{0.896 $\pm$ 0.029$^{**}$} & 0.868 $\pm$ 0.028 & \underline{0.896 $\pm$ 0.029$^{**}$} & \textbf{0.904 $\pm$ 0.031$^{***}$} \\ 
 & DeepConvNet & 0.870 $\pm$ 0.043 & 0.822 $\pm$ 0.045 & 0.833 $\pm$ 0.030 & 0.796 $\pm$ 0.064 & \underline{0.916 $\pm$ 0.031$^{***}$} & 0.916 $\pm$ 0.030$^{***}$ & 0.878 $\pm$ 0.029 & 0.916 $\pm$ 0.031$^{***}$ & \textbf{0.924 $\pm$ 0.032$^{***}$} \\ 
\midrule
\multirow{4}{*}{\makecell[l]{P300\\(Huebner2017)}} & EEGNetv4 & 0.910 $\pm$ 0.052 & 0.944 $\pm$ 0.039$^{***}$ & 0.919 $\pm$ 0.043 & 0.955 $\pm$ 0.034$^{***}$ & 0.960 $\pm$ 0.034$^{***}$ & 0.965 $\pm$ 0.029$^{***}$ & 0.953 $\pm$ 0.034$^{***}$ & \textbf{0.966 $\pm$ 0.030$^{***}$} & \underline{0.966 $\pm$ 0.029$^{***}$} \\ 
 & ATCNet & 0.903 $\pm$ 0.051 & 0.946 $\pm$ 0.040$^{***}$ & 0.908 $\pm$ 0.047 & 0.955 $\pm$ 0.036$^{***}$ & 0.963 $\pm$ 0.032$^{***}$ & \textbf{0.968 $\pm$ 0.027$^{***}$} & 0.919 $\pm$ 0.042$^{**}$ & \textbf{0.968 $\pm$ 0.028$^{***}$} & \underline{0.968 $\pm$ 0.027$^{***}$} \\ 
 & ShallowConvNet & 0.894 $\pm$ 0.056 & 0.922 $\pm$ 0.053$^{***}$ & 0.899 $\pm$ 0.052 & 0.940 $\pm$ 0.045$^{***}$ & 0.957 $\pm$ 0.037$^{***}$ & \underline{0.963 $\pm$ 0.031$^{***}$} & 0.949 $\pm$ 0.038$^{***}$ & \textbf{0.963 $\pm$ 0.031$^{***}$} & 0.962 $\pm$ 0.032$^{***}$ \\ 
 & DeepConvNet & 0.898 $\pm$ 0.053 & 0.932 $\pm$ 0.034$^{**}$ & 0.910 $\pm$ 0.042 & 0.933 $\pm$ 0.033$^{**}$ & 0.962 $\pm$ 0.032$^{***}$ & \textbf{0.964 $\pm$ 0.029$^{***}$} & 0.948 $\pm$ 0.033$^{***}$ & \underline{0.964 $\pm$ 0.029$^{***}$} & 0.964 $\pm$ 0.029$^{***}$ \\ 
\midrule
\multirow{4}{*}{\makecell[l]{P300\\(Huebner2018)}} & EEGNetv4 & 0.878 $\pm$ 0.061 & 0.938 $\pm$ 0.028$^{***}$ & 0.879 $\pm$ 0.057 & 0.951 $\pm$ 0.021$^{***}$ & 0.956 $\pm$ 0.023$^{***}$ & 0.959 $\pm$ 0.020$^{***}$ & 0.944 $\pm$ 0.028$^{***}$ & \textbf{0.960 $\pm$ 0.020$^{***}$} & \underline{0.960 $\pm$ 0.020$^{***}$} \\ 
 & ATCNet & 0.871 $\pm$ 0.067 & 0.942 $\pm$ 0.025$^{***}$ & 0.860 $\pm$ 0.059 & 0.950 $\pm$ 0.022$^{***}$ & 0.957 $\pm$ 0.022$^{***}$ & \underline{0.963 $\pm$ 0.019$^{***}$} & 0.884 $\pm$ 0.059 & \underline{0.963 $\pm$ 0.018$^{***}$} & \textbf{0.963 $\pm$ 0.018$^{***}$} \\ 
 & ShallowConvNet & 0.862 $\pm$ 0.066 & 0.914 $\pm$ 0.035$^{***}$ & 0.870 $\pm$ 0.060 & 0.935 $\pm$ 0.027$^{***}$ & 0.948 $\pm$ 0.025$^{***}$ & \underline{0.957 $\pm$ 0.021$^{***}$} & 0.937 $\pm$ 0.030$^{***}$ & \textbf{0.957 $\pm$ 0.021$^{***}$} & 0.957 $\pm$ 0.021$^{***}$ \\ 
 & DeepConvNet & 0.872 $\pm$ 0.067 & 0.936 $\pm$ 0.025$^{**}$ & 0.876 $\pm$ 0.061 & 0.939 $\pm$ 0.021$^{**}$ & 0.956 $\pm$ 0.021$^{***}$ & \underline{0.959 $\pm$ 0.021$^{***}$} & 0.940 $\pm$ 0.028$^{***}$ & \textbf{0.959 $\pm$ 0.021$^{***}$} & 0.959 $\pm$ 0.020$^{***}$ \\ 
\midrule
\multirow{4}{*}{\makecell[l]{SSVEP\\(Lee2019)}} & EEGNetv4 & 0.931 $\pm$ 0.117 & 0.663 $\pm$ 0.145 & 0.935 $\pm$ 0.119 & 0.680 $\pm$ 0.108 & 0.953 $\pm$ 0.089$^{***}$ & 0.955 $\pm$ 0.084$^{***}$ & 0.946 $\pm$ 0.106$^{**}$ & \underline{0.956 $\pm$ 0.084$^{***}$} & \textbf{0.956 $\pm$ 0.088$^{***}$} \\ 
 & ATCNet & 0.933 $\pm$ 0.110 & 0.497 $\pm$ 0.151 & 0.944 $\pm$ 0.102 & 0.464 $\pm$ 0.097 & 0.953 $\pm$ 0.089$^{**}$ & \underline{0.958 $\pm$ 0.079$^{**}$} & 0.946 $\pm$ 0.097$^{*}$ & \textbf{0.958 $\pm$ 0.080$^{**}$} & 0.955 $\pm$ 0.083$^{**}$ \\ 
 & ShallowConvNet & 0.871 $\pm$ 0.140 & 0.376 $\pm$ 0.125 & 0.886 $\pm$ 0.141$^{*}$ & 0.325 $\pm$ 0.062 & 0.921 $\pm$ 0.099$^{***}$ & 0.924 $\pm$ 0.099$^{***}$ & 0.913 $\pm$ 0.111$^{***}$ & \underline{0.924 $\pm$ 0.095$^{***}$} & \textbf{0.928 $\pm$ 0.096$^{***}$} \\ 
 & DeepConvNet & 0.948 $\pm$ 0.108 & 0.319 $\pm$ 0.091 & 0.951 $\pm$ 0.106 & 0.312 $\pm$ 0.069 & 0.961 $\pm$ 0.087$^{**}$ & \underline{0.966 $\pm$ 0.082$^{**}$} & 0.959 $\pm$ 0.090$^{*}$ & 0.965 $\pm$ 0.084$^{**}$ & \textbf{0.966 $\pm$ 0.082$^{**}$} \\ 
\midrule
\multirow{4}{*}{\makecell[l]{SSVEP\\(Kalunga2016)}} & EEGNetv4 & 0.352 $\pm$ 0.153 & 0.240 $\pm$ 0.049 & 0.365 $\pm$ 0.136 & 0.242 $\pm$ 0.051 & 0.362 $\pm$ 0.178 & \underline{0.380 $\pm$ 0.149} & 0.378 $\pm$ 0.133 & 0.379 $\pm$ 0.151 & \textbf{0.383 $\pm$ 0.146} \\ 
 & ATCNet & 0.514 $\pm$ 0.147 & 0.233 $\pm$ 0.064 & 0.442 $\pm$ 0.180 & 0.236 $\pm$ 0.067 & \textbf{0.547 $\pm$ 0.152} & \underline{0.544 $\pm$ 0.155} & 0.511 $\pm$ 0.160 & 0.541 $\pm$ 0.157 & 0.535 $\pm$ 0.142 \\ 
 & ShallowConvNet & 0.409 $\pm$ 0.153 & 0.292 $\pm$ 0.059 & 0.433 $\pm$ 0.152 & 0.294 $\pm$ 0.088 & \textbf{0.474 $\pm$ 0.136} & \underline{0.465 $\pm$ 0.172} & 0.424 $\pm$ 0.147 & \underline{0.465 $\pm$ 0.172} & 0.453 $\pm$ 0.157 \\ 
 & DeepConvNet & 0.291 $\pm$ 0.042 & 0.255 $\pm$ 0.033 & \textbf{0.304 $\pm$ 0.073} & 0.238 $\pm$ 0.032 & \underline{0.298 $\pm$ 0.052} & 0.260 $\pm$ 0.077 & 0.254 $\pm$ 0.083 & 0.268 $\pm$ 0.088 & 0.291 $\pm$ 0.072 \\ 
\midrule
\multirow{4}{*}{\makecell[l]{SSVEP\\(MAMEM2)}} & EEGNetv4 & 0.210 $\pm$ 0.032 & 0.204 $\pm$ 0.032 & 0.216 $\pm$ 0.047 & 0.202 $\pm$ 0.016 & \textbf{0.224 $\pm$ 0.033} & \underline{0.218 $\pm$ 0.029} & 0.204 $\pm$ 0.026 & 0.211 $\pm$ 0.031 & 0.202 $\pm$ 0.030 \\ 
 & ATCNet & 0.193 $\pm$ 0.025 & \textbf{0.233 $\pm$ 0.016$^{*}$} & 0.214 $\pm$ 0.037 & 0.206 $\pm$ 0.026 & \underline{0.230 $\pm$ 0.037} & 0.224 $\pm$ 0.050 & 0.208 $\pm$ 0.046 & 0.207 $\pm$ 0.047 & 0.209 $\pm$ 0.036 \\ 
 & ShallowConvNet & 0.186 $\pm$ 0.040 & 0.214 $\pm$ 0.038 & 0.201 $\pm$ 0.033 & 0.194 $\pm$ 0.038 & 0.211 $\pm$ 0.026 & 0.200 $\pm$ 0.025 & \textbf{0.227 $\pm$ 0.038} & 0.200 $\pm$ 0.025 & \underline{0.219 $\pm$ 0.047} \\ 
 & DeepConvNet & \textbf{0.217 $\pm$ 0.050} & 0.203 $\pm$ 0.023 & 0.202 $\pm$ 0.029 & 0.186 $\pm$ 0.026 & \underline{0.210 $\pm$ 0.034} & 0.193 $\pm$ 0.036 & 0.191 $\pm$ 0.038 & 0.198 $\pm$ 0.047 & 0.197 $\pm$ 0.035 \\ 
\bottomrule
\end{tabular}%
}
\label{tab:full_adaptation_comparison}
\end{table}
\FloatBarrier

\subsection{Trial-by-trial learning dynamics and per-subject performance}

\begin{figure*}[htbp!]
\centering
\includegraphics[width=\textwidth]{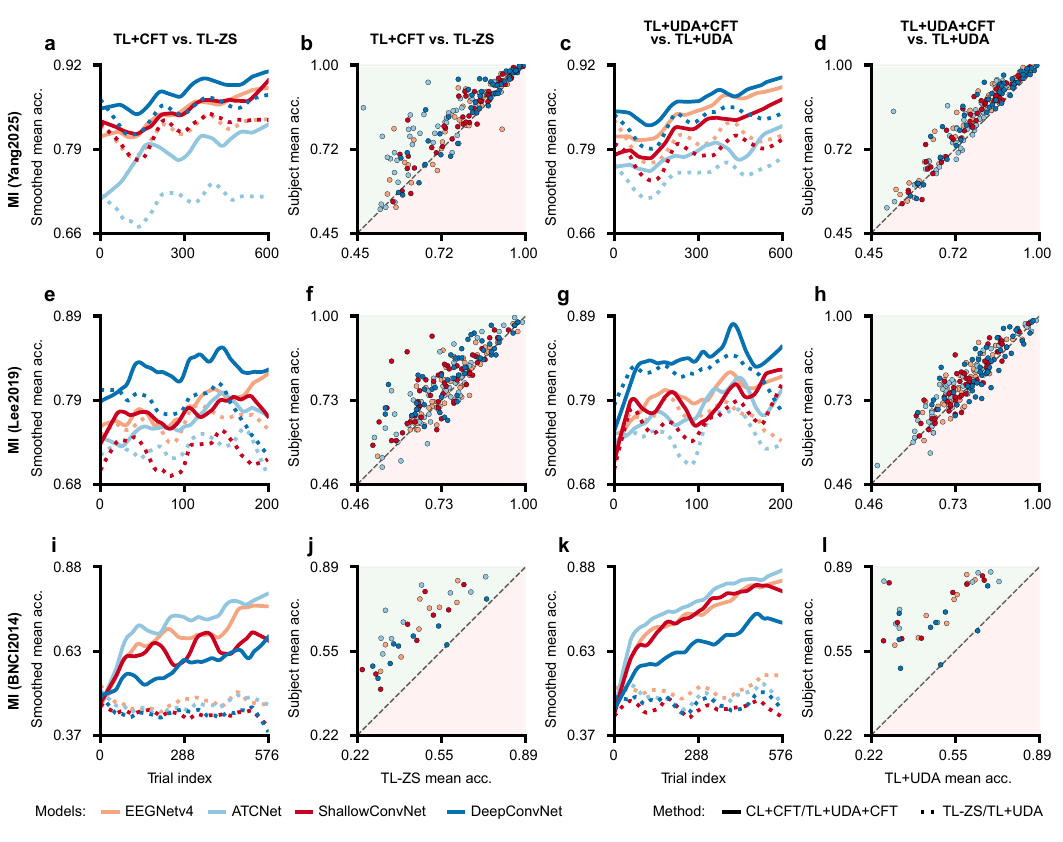}
\caption{\textbf{Trial-by-trial and per-subject effect of continual finetuning across Motor Imagery (MI) datasets.} Each row represents a single dataset. The columns are organized into two comparison groups, each with two types of plots. Columns 1-2 compare a PRE+CFT model to a static PRE-ZS baseline. Columns 3-4 compare the PRE+UDA+CFT model to a PRE+UDA baseline. Plot Types: The first column in each comparison (1 and 3) is a time-series plot showing smoothed accuracy over trials (solid lines: +CFT; dotted lines: no CFT). The second column (2 and 4) is a per-subject scatter plot comparing mean accuracy with CFT (y-axis) against the corresponding baseline without it (x-axis). Points above the diagonal signify performance improvement for an individual subject due to CFT.
} 
\label{figs1a}
\end{figure*}

\begin{figure*}[htbp!]
\centering
\includegraphics[width=\textwidth]{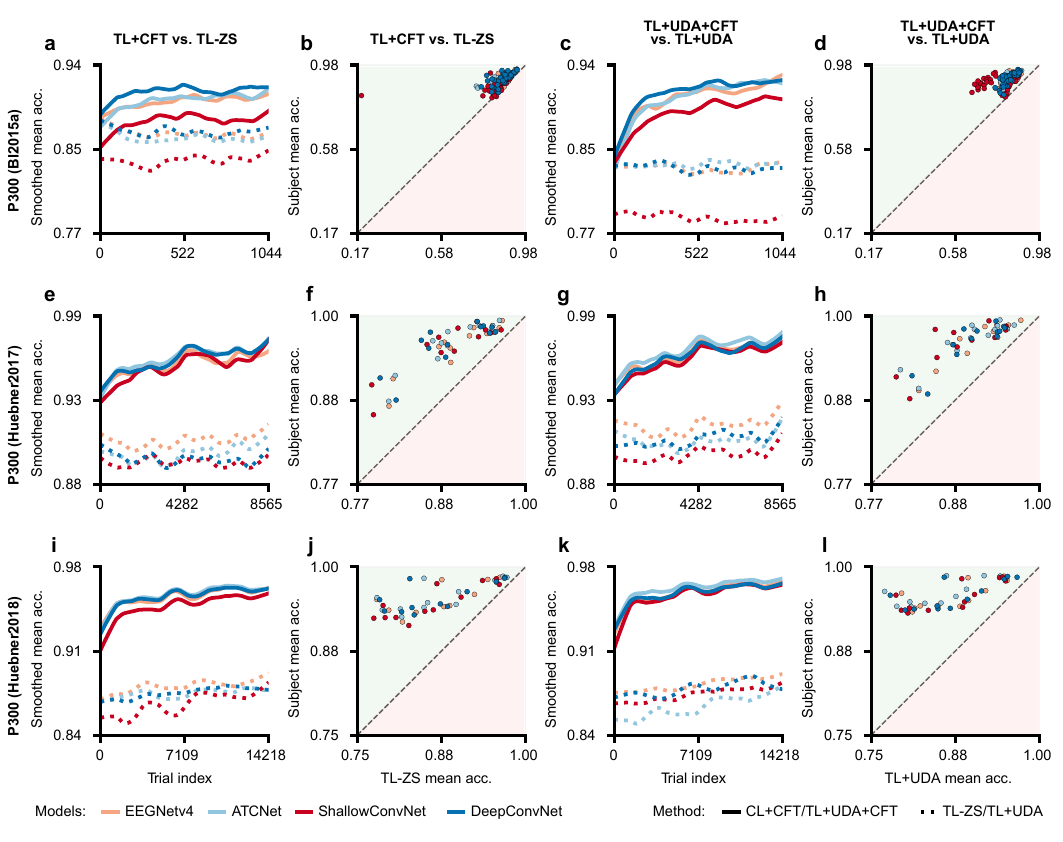}
\caption{\textbf{Trial-by-trial and per-subject effect of continual finetuning across P300 datasets.} Each row represents a single dataset. The columns are organized into two comparison groups, each with two types of plots. Columns 1-2 compare a PRE+CFT model to a static PRE-ZS baseline. Columns 3-4 compare the PRE+UDA+CFT model to a PRE+UDA baseline. Plot Types: The first column in each comparison (1 and 3) is a time-series plot showing smoothed accuracy over trials (solid lines: +CFT; dotted lines: no CFT). The second column (2 and 4) is a per-subject scatter plot comparing mean accuracy with CFT (y-axis) against the corresponding baseline without it (x-axis). Points above the diagonal signify performance improvement for an individual subject due to CFT.}
\label{figs1b}
\end{figure*}

\begin{figure*}[htbp!]
\centering
\includegraphics[width=\textwidth]{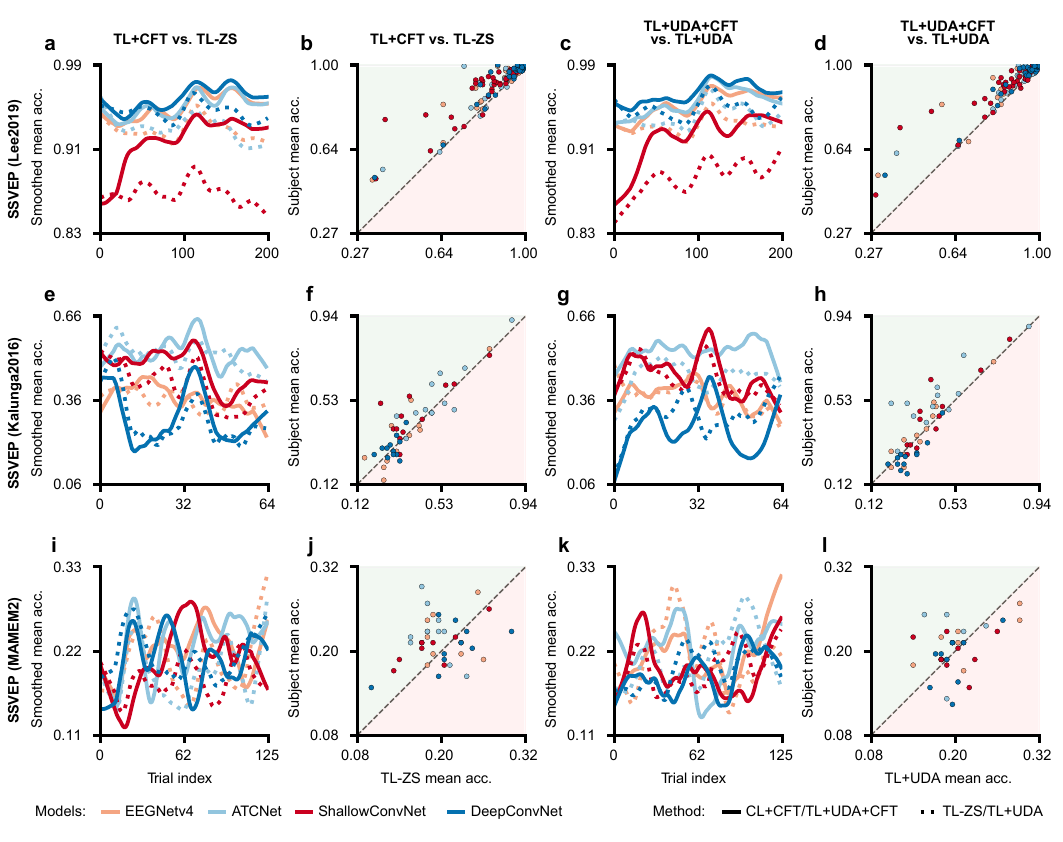}
\caption{\textbf{Trial-by-trial and per-subject effect of continual finetuning across SSVEP datasets.} Each row represents a single dataset. The columns are organized into two comparison groups, each with two types of plots. Columns 1-2 compare a PRE+CFT model to a static PRE-ZS baseline. Columns 3-4 compare the PRE+UDA+CFT model to a PRE+UDA baseline. Plot Types: The first column in each comparison (1 and 3) is a time-series plot showing smoothed accuracy over trials (solid lines: +CFT; dotted lines: no CFT). The second column (2 and 4) is a per-subject scatter plot comparing mean accuracy with CFT (y-axis) against the corresponding baseline without it (x-axis). Points above the diagonal signify performance improvement for an individual subject due to CFT.}
\label{figs1c}
\end{figure*}

\begin{figure*}[htbp!]
\centering
\includegraphics[width=\textwidth]{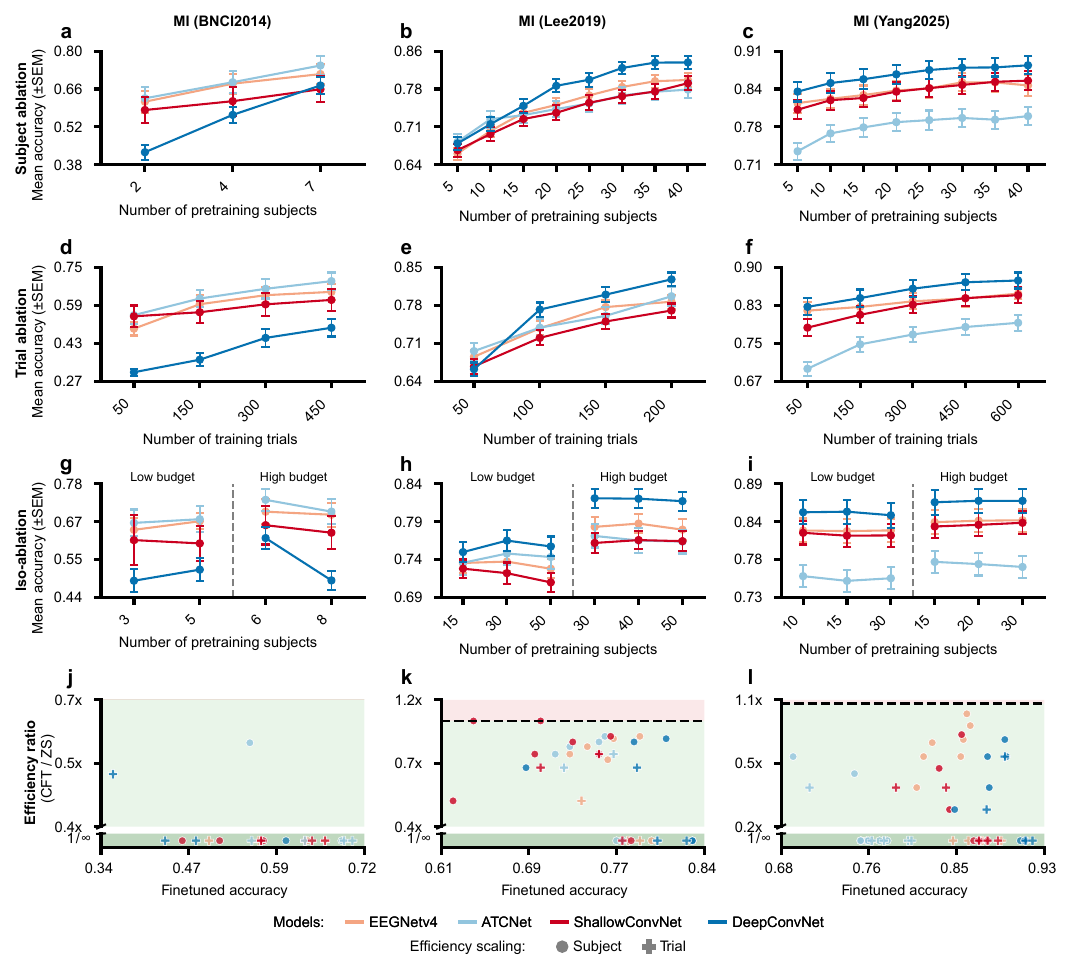}
\caption{\textbf{Scaling performance of continually finetuned models on Motor Imagery (MI) datasets.} The plots show the mean classification accuracy (±SEM) for four deep learning models after applying continual finetuning (CFT). Each column corresponds to a different MI dataset: BNCI2014_001, Lee2019_MI, and Yang2025. \textbf{(a-c)} Subject ablation: Performance as a function of the number of pretraining subjects. \textbf{(d-f)} Trial ablation: Performance as a function of the number of trials per pretraining subject. \textbf{(g-i)} Iso-ablation: Performance under fixed "Low" and "High" total pretraining trial budgets. For BNCI2014_001, the budgets are  $\approx$1.5k and  $\approx$3k trials; for Lee2019_MI,  $\approx$3k and  $\approx$6k trials; and for Yang2025,  $\approx$5k and  $\approx$10k trials. The plots illustrate the trade-off between fewer subjects with more trials each (left within each budget group) versus more subjects with fewer trials each (right).
\textbf{(j-l): CFT efficiency gains}. For a fixed accuracy (within 1\% absolute tolerance), pairs of PRE+CFT and PRE-ZS configurations are chosen from the scaling plots, and ratio of pretraining data used by the two is plotted against the achieved accuracy. In most cases, CFT required less data to achieve the same accuracy and sometimes outperforms ZS accuracies using full data.}
\label{figs2a}
\end{figure*}

\begin{figure*}[htbp!]
\centering
\includegraphics[width=\textwidth]{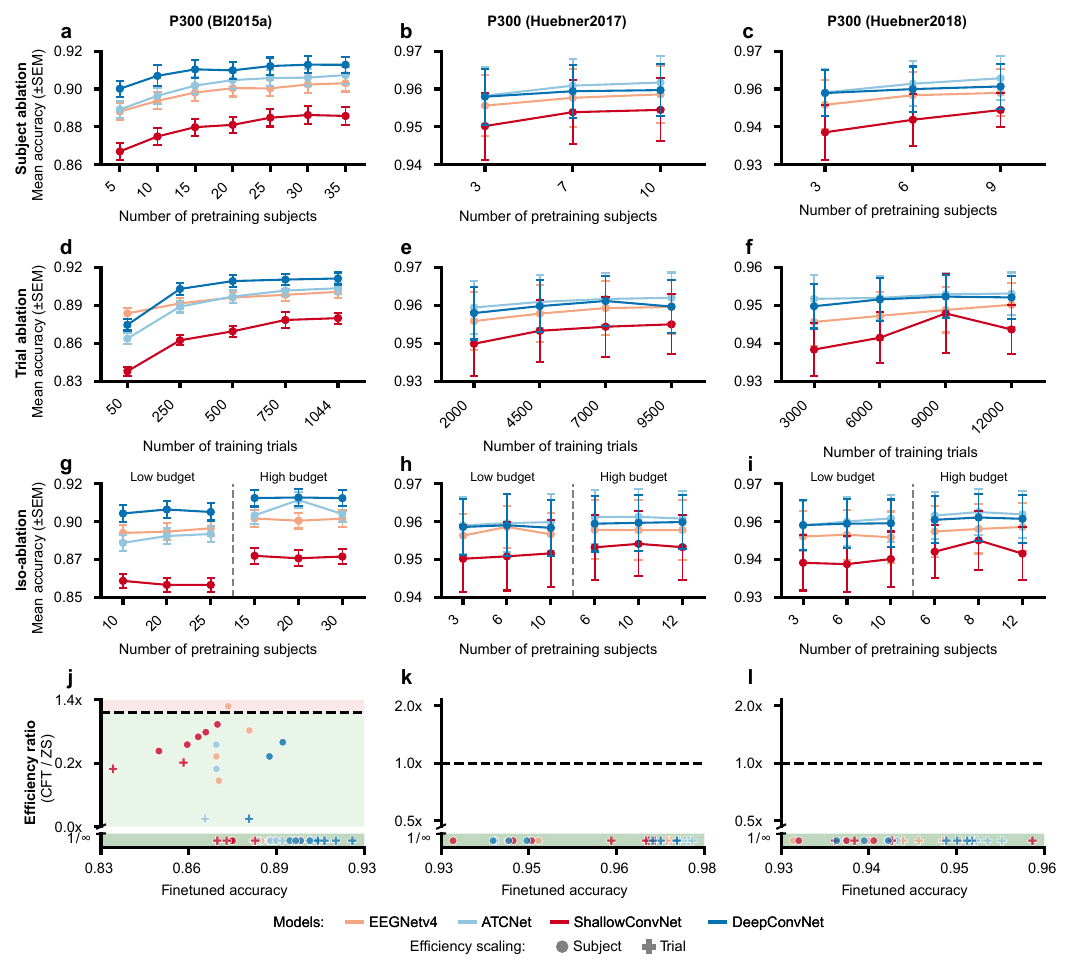}
\caption{\textbf{Scaling performance of continually finetuned models on P300 datasets.} The plots show the mean classification accuracy (±SEM) for four deep learning models after applying continual finetuning (CFT). Each column corresponds to a different P300 dataset: BI2015a, Huebner2017, and Huebner2018. \textbf{(a-c)} Subject ablation: Performance as a function of the number of pretraining subjects. \textbf{(d-f)} Trial ablation: Performance as a function of the number of trials per pretraining subject. \textbf{(g-i)} Iso-ablation: Performance under fixed "Low" and "High" total pretraining trial budgets. For BI2015a, the budgets are  $\approx$10k and  $\approx$20k trials; for Huebner2017,  $\approx$37.5k and  $\approx$75k trials; and for Huebner2018,  $\approx$42k and  $\approx$84k trials. The plots illustrate the trade-off between fewer subjects with more trials each (left within each budget group) versus more subjects with fewer trials each (right).
\textbf{(j-l): CFT efficiency gains}. For a fixed accuracy (within 1\% absolute tolerance), pairs of PRE+CFT and PRE-ZS configurations are chosen from the scaling plots, and ratio of pretraining data used by the two is plotted against the achieved accuracy. CFT required less data to achieve the same accuracy, and in two datasets fully outperforms ZS accuracies using full data.}
\label{figs2b}
\end{figure*}

\begin{figure*}[htbp!]
\centering
\includegraphics[width=\textwidth]{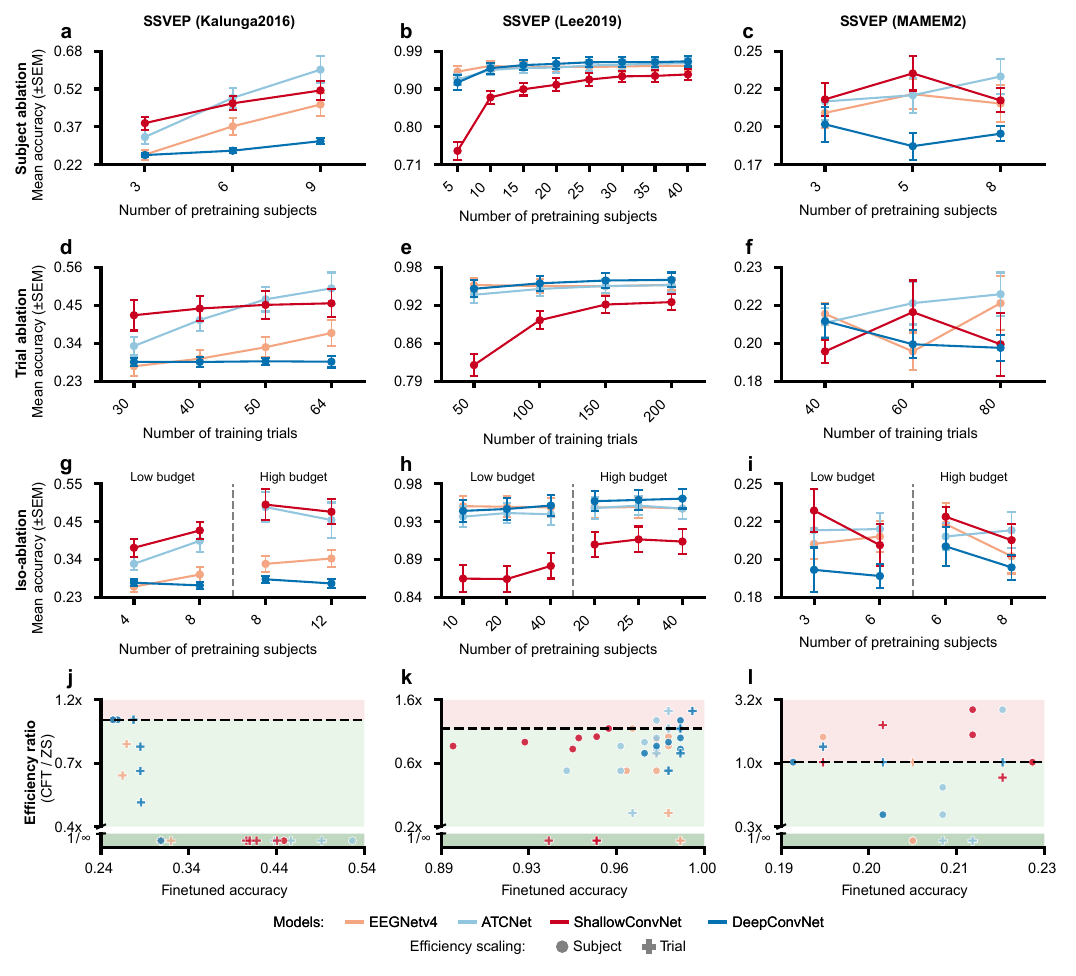}
\caption{\textbf{Scaling performance of continually finetuned models on SSVEP datasets.} The plots show the mean classification accuracy (±SEM) for four deep learning models after applying continual finetuning (CFT). Each column corresponds to a different SSVEP dataset: Kalunga2016, Lee2019_SSVEP, and MAMEM2. \textbf{(a-c)} Subject ablation: Performance as a function of the number of pretraining subjects. \textbf{(d-f)} Trial ablation: Performance as a function of the number of trials per pretraining subject. \textbf{(g-i)} Iso-ablation: Performance under fixed "Low" and "High" total pretraining trial budgets. For Kalunga2016, the budgets are  $\approx$256 and  $\approx$512 trials; for Lee2019_SSVEP,  $\approx$2k and  $\approx$4k trials; and for MAMEM2,  $\approx$300 and  $\approx$600 trials. The plots illustrate the trade-off between fewer subjects with more trials each (left within each budget group) versus more subjects with fewer trials each (right).}
\label{figs2c}
\end{figure*}

\begin{figure*}[htbp!]
\centering
\includegraphics[width=\textwidth]{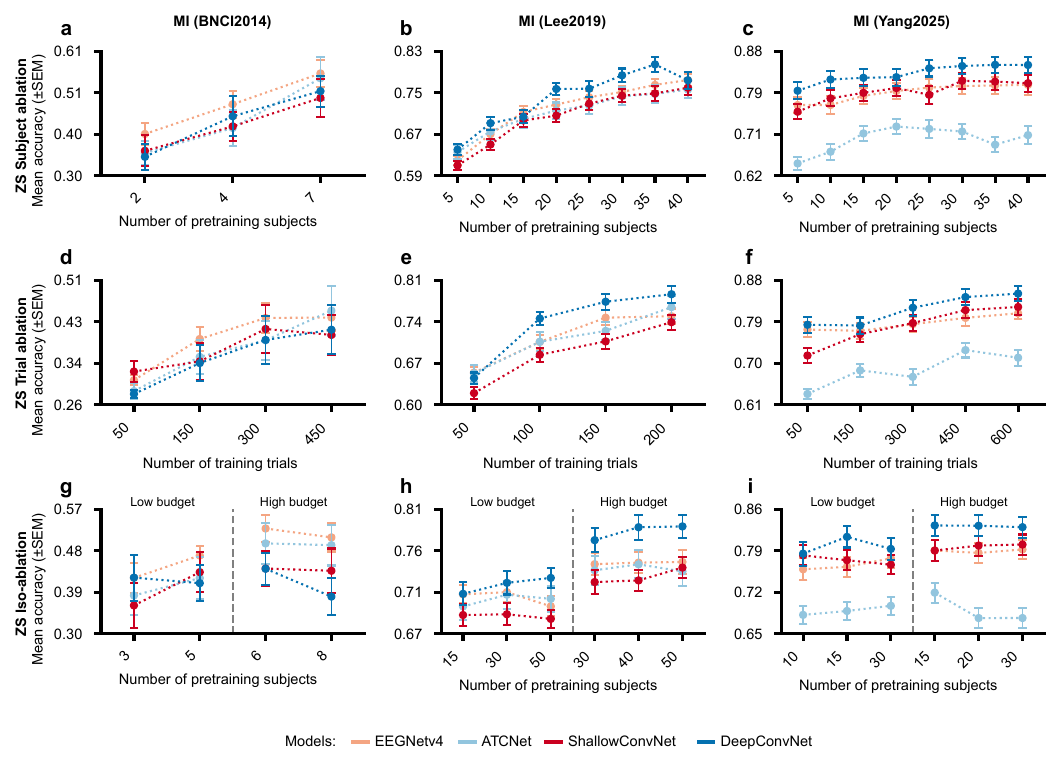}
\caption{\textbf{Scaling performance of zero-shot population-level pretraining models on Motor Imagery (MI) datasets.} The plots show the mean classification accuracy (±SEM) for four deep learning models evaluated in a zero-shot (ZS) setting. Each column corresponds to a different MI dataset: BNCI2014_001, Lee2019_MI, and Yang2025. \textbf{(a-c)} Subject ablation: Performance as a function of the number of pretraining subjects. \textbf{(d-f)} Trial ablation: Performance as a function of the number of trials per pretraining subject. \textbf{(g-i)} Iso-ablation: Performance under fixed "Low" and "High" total pretraining trial budgets. For BNCI2014_001, the budgets are  $\approx$1.5k and  $\approx$3k trials; for Lee2019_MI,  $\approx$3k and  $\approx$6k trials; and for Yang2025,  $\approx$5k and  $\approx$10k trials, revealing the pretraining data trade-offs.
\textbf{(j-l): CFT efficiency gains}. For a fixed accuracy (within 1\% absolute tolerance), pairs of PRE+CFT and PRE-ZS configurations are chosen from the scaling plots, and ratio of pretraining data used by the two is plotted against the achieved accuracy. In several cases, CFT required less data to achieve the same accuracy and sometimes outperforms ZS accuracies using full data.}
\label{figs2d}
\end{figure*}

\begin{figure*}[htbp!]
\centering
\includegraphics[width=\textwidth]{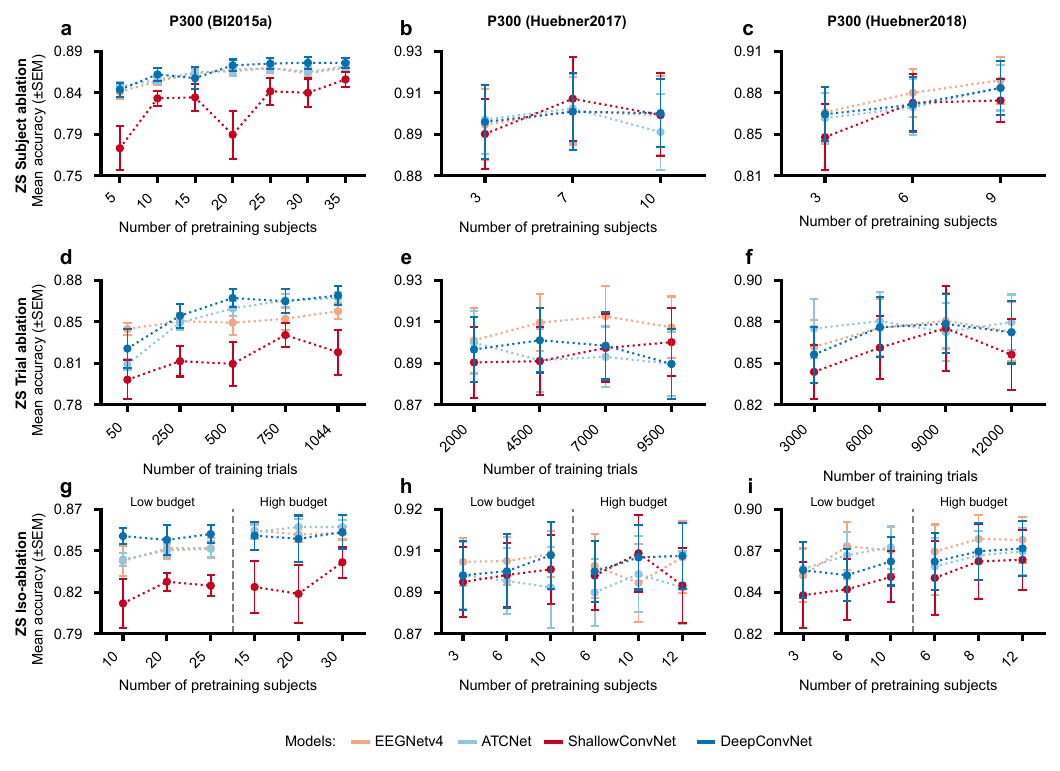}
\caption{\textbf{Scaling performance of zero-shot population-level pretraining models on P300 datasets.} The plots show the mean classification accuracy (±SEM) for four deep learning models evaluated in a zero-shot (ZS) setting. Each column corresponds to a different P300 dataset: BI2015a, Huebner2017, and Huebner2018. \textbf{(a-c)} Subject ablation: Performance as a function of the number of pretraining subjects. \textbf{(d-f)} Trial ablation: Performance as a function of the number of trials per pretraining subject. \textbf{(g-i)} Iso-ablation: Performance under fixed "Low" and "High" total pretraining trial budgets. For BI2015a, the budgets are  $\approx$10k and  $\approx$20k trials; for Huebner2017,  $\approx$37.5k and  $\approx$75k trials; and for Huebner2018,  $\approx$42k and  $\approx$84k trials, revealing the pretraining data trade-offs.}
\label{figs2e}
\end{figure*}

\begin{figure*}[htbp!]
\centering
\includegraphics[width=\textwidth]{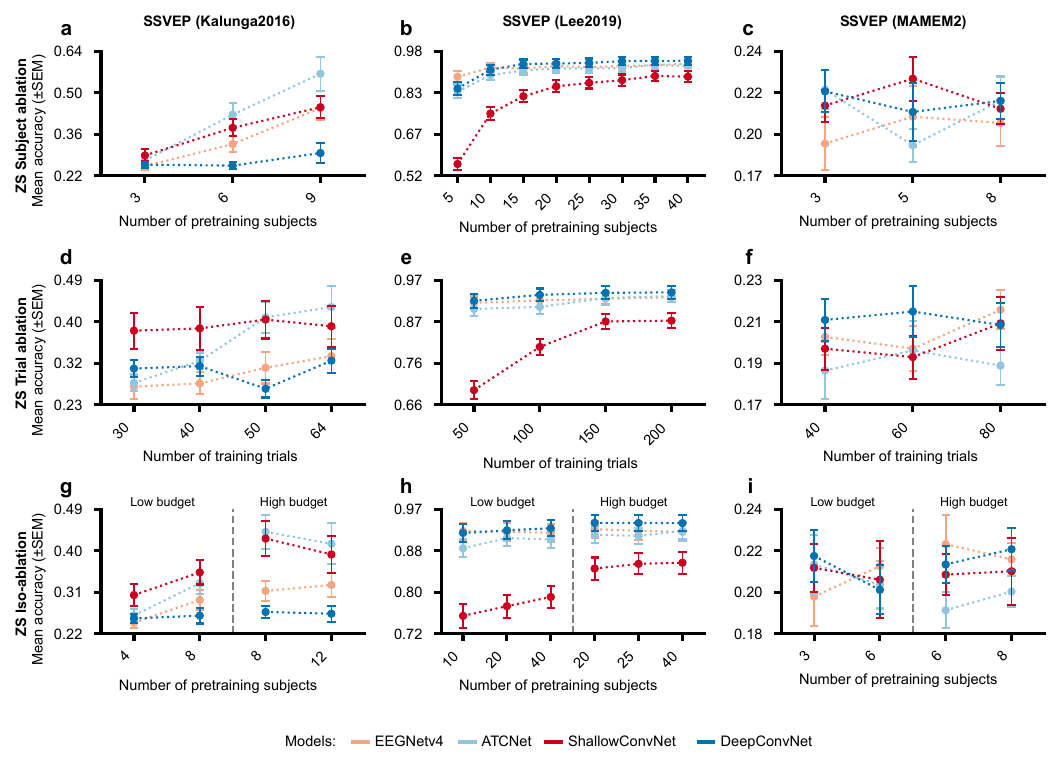}
\caption{\textbf{Scaling performance of zero-shot population-level pretraining models on SSVEP datasets.} The plots show the mean classification accuracy (±SEM) for four deep learning models evaluated in a zero-shot (ZS) setting. Each column corresponds to a different SSVEP dataset: Kalunga2016, Lee2019_SSVEP, and MAMEM2. \textbf{(a-c)} Subject ablation: Performance as a function of the number of pretraining subjects. \textbf{(d-f)} Trial ablation: Performance as a function of the number of trials per pretraining subject. \textbf{(g-i)} Iso-ablation: Performance under fixed "Low" and "High" total pretraining trial budgets. For Kalunga2016, the budgets are  $\approx$256 and  $\approx$512 trials; for Lee2019_SSVEP,  $\approx$2k and  $\approx$4k trials; and for MAMEM2,  $\approx$300 and  $\approx$600 trials, revealing the pretraining data trade-offs.}
\label{figs2f}
\end{figure*}

We present the complete visualizations of trial-by-trial learning dynamics and per-subject performance gains for all nine datasets, which support the findings in the main text (Fig. \ref{figs1a}-\ref{figs1c}). .

\FloatBarrier

\FloatBarrier

\FloatBarrier

\subsection{Computational performance and real-time feasibility}

To validate the suitability of the EDAPT framework for practical, real-time BCI applications, we conducted a comprehensive analysis of its computational latency. We benchmarked the three core online components—the optional unsupervised domain adaptation (UDA) statistics update, prediction (forward pass), and a single-batch continual finetuning (CFT) update—across all nine datasets and four deep learning architectures. The performance was measured on both a standard multi-core CPU and a consumer-grade GPU to provide a clear overview of the framework's operational efficiency under different hardware scenarios (Table \ref{tab:latency_summary_cpu_gpu}). 

\begin{table}[htbp]
\centering
\caption{\textbf{Latency benchmarks for EDAPT's online components.} Median computational time in milliseconds (ms) for the three online stages of the EDAPT framework: prediction (a single forward pass), the unsupervised domain adaptation (UDA) update, and one continual finetuning (CFT) epoch. Latencies were measured for all nine datasets and four network architectures, benchmarked on an 8-core CPU  and a consumer-grade GPU (NVIDIA GeForce RTX 2080Ti, 11 GB). The results demonstrate that the total time for the critical real-time path (UDA update + prediction) is consistently low, particularly on the GPU, underscoring the framework's suitability for interactive BCI. The more intensive CFT step shows feasibility for inter-trial updates with GPU acceleration. The highest latencies were observed for the MAMEM2 dataset, which is expected given its high channel count (256 channels)}
\resizebox{\textwidth}{!}{
\begin{tabular}{lrr l rrrrrr} 
\toprule
\multirow{2}{*}{\begin{tabular}[c]{@{}c@{}}Dataset \\ (Paradigm)\end{tabular}} & \multirow{2}{*}{\begin{tabular}[c]{@{}c@{}}Trial \\ length (s)\end{tabular}} & \multirow{2}{*}{\begin{tabular}[c]{@{}c@{}}Sampling \\ rate (Hz)\end{tabular}} & \multirow{2}{*}{Model} & \multicolumn{2}{c}{Prediction (ms)} & \multicolumn{2}{c}{UDA update (ms)} & \multicolumn{2}{c}{\begin{tabular}[c]{@{}c@{}}Finetune \\ epoch (ms)\end{tabular}} \\ 
\cmidrule(lr){5-6} \cmidrule(lr){7-8} \cmidrule(lr){9-10}
 &  &  &  & CPU & GPU & CPU & GPU & CPU & GPU \\ 
\midrule
\multirow{4}{*}{\begin{tabular}[c]{@{}c@{}}BNCI2014\_001 \\ (MI)\end{tabular}} & \multirow{4}{*}{4} & \multirow{4}{*}{250} & EEGNetv4 & 1.97 & 0.75 & 0.86 & 0.99 & 157.05 & 9.11 \\ 
 & & & ATCNet & 8.04 & 6.84 & 0.63 & 0.76 & 179.83 & 25.61 \\ 
 & & & ShallowConvNet & 1.83 & 0.95 & 0.63 & 0.77 & 266.44 & 38.87 \\ 
 & & & DeepConvNet & 2.38 & 1.07 & 0.63 & 0.99 & 225.46 & 18.66 \\ 
\midrule
\multirow{4}{*}{\begin{tabular}[c]{@{}c@{}}Lee2019\_MI \\ (MI)\end{tabular}} & \multirow{4}{*}{4} & \multirow{4}{*}{1000} & EEGNetv4 & 10.67 & 0.84 & 2.14 & 2.48 & 1318.65 & 107.23 \\ 
 & & & ATCNet & 21.30 & 7.77 & 2.83 & 3.12 & 1591.87 & 125.79 \\ 
 & & & ShallowConvNet & 20.01 & 3.20 & 2.12 & 3.13 & 4263.86 & 76.72 \\ 
 & & & DeepConvNet & 16.12 & 2.79 & 2.12 & 2.46 & 2457.56 & 205.66 \\ 
\midrule
\multirow{4}{*}{\begin{tabular}[c]{@{}c@{}}Yang2025 \\ (MI)\end{tabular}} & \multirow{4}{*}{4} & \multirow{4}{*}{1000} & EEGNetv4 & 9.71 & 0.85 & 1.90 & 2.85 & 1234.18 & 101.66 \\ 
 & & & ATCNet & 16.55 & 6.60 & 1.88 & 2.23 & 1544.55 & 110.53 \\ 
 & & & ShallowConvNet & 21.04 & 3.18 & 2.55 & 2.84 & 4092.96 & 72.80 \\ 
 & & & DeepConvNet & 17.30 & 2.72 & 2.55 & 2.24 & 2618.35 & 185.05 \\ 
\midrule
\multirow{4}{*}{\begin{tabular}[c]{@{}c@{}}BI2015a \\ (P300)\end{tabular}} & \multirow{4}{*}{1} & \multirow{4}{*}{512} & EEGNetv4 & 1.17 & 0.76 & 0.75 & 1.14 & 104.10 & 7.63 \\ 
 & & & ATCNet & 7.38 & 6.87 & 0.76 & 0.87 & 134.27 & 24.93 \\ 
 & & & ShallowConvNet & 1.78 & 0.65 & 1.01 & 0.88 & 222.83 & 20.28 \\ 
 & & & DeepConvNet & 2.44 & 1.00 & 1.01 & 1.15 & 182.76 & 14.95 \\ 
\midrule
\multirow{4}{*}{\begin{tabular}[c]{@{}c@{}}Huebner2017 \\ (P300)\end{tabular}} & \multirow{4}{*}{0.9} & \multirow{4}{*}{1000} & EEGNetv4 & 2.08 & 0.67 & 1.06 & 0.90 & 195.96 & 12.46 \\ 
 & & & ATCNet & 9.97 & 6.90 & 1.06 & 0.91 & 240.38 & 28.32 \\ 
 & & & ShallowConvNet & 1.91 & 0.93 & 0.79 & 1.20 & 325.85 & 37.96 \\ 
 & & & DeepConvNet & 2.93 & 1.08 & 0.82 & 1.20 & 278.17 & 23.76 \\ 
\midrule
\multirow{4}{*}{\begin{tabular}[c]{@{}c@{}}Huebner2018 \\ (P300)\end{tabular}} & \multirow{4}{*}{0.9} & \multirow{4}{*}{1000} & EEGNetv4 & 2.08 & 0.68 & 1.07 & 0.91 & 193.37 & 12.91 \\ 
 & & & ATCNet & 7.99 & 7.79 & 0.80 & 1.22 & 204.21 & 30.60 \\ 
 & & & ShallowConvNet & 1.92 & 0.91 & 0.80 & 0.92 & 326.85 & 37.91 \\ 
 & & & DeepConvNet & 2.90 & 1.05 & 0.80 & 0.92 & 281.95 & 23.78 \\ 
\midrule
\multirow{4}{*}{\begin{tabular}[c]{@{}c@{}}Kalunga2016 \\ (SSVEP)\end{tabular}} & \multirow{4}{*}{2} & \multirow{4}{*}{256} & EEGNetv4 & 1.31 & 0.73 & 0.52 & 0.64 & 45.62 & 3.28 \\ 
 & & & ATCNet & 9.01 & 6.63 & 0.57 & 0.53 & 86.97 & 22.05 \\ 
 & & & ShallowConvNet & 0.85 & 0.44 & 0.43 & 0.54 & 47.88 & 4.14 \\ 
 & & & DeepConvNet & 1.50 & 0.76 & 0.43 & 0.53 & 52.00 & 8.03 \\ 
\midrule
\multirow{4}{*}{\begin{tabular}[c]{@{}c@{}}Lee2019\_SSVEP \\ (SSVEP)\end{tabular}} & \multirow{4}{*}{4} & \multirow{4}{*}{1000} & EEGNetv4 & 10.18 & 0.88 & 2.09 & 3.09 & 1317.18 & 106.78 \\ 
 & & & ATCNet & 17.01 & 6.58 & 2.09 & 2.45 & 1366.33 & 117.87 \\ 
 & & & ShallowConvNet & 21.75 & 3.16 & 2.79 & 2.44 & 4677.09 & 75.15 \\ 
 & & & DeepConvNet & 17.29 & 2.17 & 2.80 & 2.44 & 2708.15 & 206.75 \\ 
\midrule
\multirow{4}{*}{\begin{tabular}[c]{@{}c@{}}MAMEM2 \\ (SSVEP)\end{tabular}} & \multirow{4}{*}{3} & \multirow{4}{*}{250} & EEGNetv4 & 7.94 & 0.81 & 166.91 & 143.03 & 1046.16 & 221.30 \\ 
 & & & ATCNet & 18.09 & 7.63 & 155.77 & 154.80 & 903.79 & 229.86 \\ 
 & & & ShallowConvNet & 15.69 & 2.44 & 154.26 & 159.81 & 2276.15 & 77.76 \\ 
 & & & DeepConvNet & 7.29 & 1.90 & 147.25 & 151.57 & 1658.95 & 515.88 \\ 
\bottomrule
\end{tabular}%
}
\label{tab:latency_summary_cpu_gpu}
\end{table}
\FloatBarrier

\subsection{Scaling analysis of pretraining data}
To provide a complete picture of how pretraining data composition affects downstream BCI performance, we present the full results of our scaling analyses for all nine datasets. These experiments systematically evaluate the impact of varying the number of pretraining subjects (Subject ablation), the number of trials per pretraining subject (Trial ablation), and the trade-off between these two factors under a fixed data budget (Iso-ablation). The results are provided for models with subsequent continual finetuning (PRE+CFT), which demonstrates the effect on the full EDAPT framework, and for zero-shot (PRE-ZS) models, which isolates the effect of pretraining alone. These figures support the conclusions drawn in the main text, showing a consistent logarithmic scaling with data and that model performance is primarily driven by the total pretraining data budget, not its specific composition (i.e., the trade-off between more subjects or more trials).

\begin{table}[h]
\centering
\caption{Configuration levels for the subject ablation scaling analysis. For each dataset, the number of pretraining subjects was varied as listed, while the number of trials per subject was held constant at the maximum available for that dataset.}
\label{tab:subject_ablation_configs}
\footnotesize %
\begin{tabularx}{\textwidth}{@{}>{\raggedright\arraybackslash}X >{\raggedright\arraybackslash}X >{\centering\arraybackslash}X >{\raggedright\arraybackslash}X@{}}
\toprule
\textbf{Paradigm} & \textbf{Dataset} & \textbf{Trials per subject (fixed)} & \textbf{Subjects (varied)} \\
\midrule
\textbf{MI} & BNCI2014\_001 & 576 & 2, 4, 6, 8 \\
& Lee2019\_MI & 200 & 5, 10, 15, 20, 25, 30, 35, 40 \\
& Yang2025 & 600 & 5, 10, 15, 20, 25, 30, 35, 40 \\
\midrule
\textbf{P300} & BI2015a & 1044 & 5, 10, 15, 20, 25, 30, 35 \\
& Huebner2017 & 12850 & 3, 6, 9, 12 \\
& Huebner2018 & 14275 & 3, 6, 9 \\
\midrule
\textbf{SSVEP} & Kalunga2016 & 64 & 3, 6, 9 \\
& Lee2019\_SSVEP & 200 & 5, 10, 15, 20, 25, 30, 35, 40 \\
& MAMEM2 & 100 & 3, 5, 8 \\
\end{tabularx}
\end{table}
\FloatBarrier

\begin{table}[h]
\centering
\caption{Configuration levels for the trial ablation scaling analysis. For each dataset, the number of trials per subject was varied as listed, while the number of subjects was held constant at the maximum available for that dataset.}
\label{tab:trial_ablation_configs}
\footnotesize %
\begin{tabularx}{\textwidth}{@{}>{\raggedright\arraybackslash}X >{\raggedright\arraybackslash}X >{\centering\arraybackslash}X >{\raggedright\arraybackslash}X@{}}
\toprule
\textbf{Paradigm} & \textbf{Dataset} & \textbf{Subjects (fixed)} & \textbf{Trials per subject (varied)} \\
\midrule
\textbf{MI} & BNCI2014\_001 & 9 & 50, 150, 300, 450 \\
& Lee2019\_MI & 54 & 50, 100, 150 \\
& Yang2025 & 51 & 50, 150, 300, 450, 600 \\
\midrule
\textbf{P300} & BI2015a & 43 & 50, 250, 500, 750, 1044 \\
& Huebner2017 & 13 & 2000, 4500, 7000, 9500 \\
& Huebner2018 & 12 & 3000, 6000, 9000, 12000 \\
\midrule
\textbf{SSVEP} & Kalunga2016 & 12 & 30, 40, 50, 64, 70, 90 \\
& Lee2019\_SSVEP & 54 & 50, 100, 150, 200 \\
& MAMEM2 & 10 & 40, 60, 80 \\
\bottomrule
\end{tabularx}
\end{table}
\FloatBarrier

\begin{table}[h]
\centering
\caption{Configuration levels for the iso-ablation scaling analysis to study the trade-off between the number of subjects and trials per subject under fixed data budgets.}
\label{tab:iso_ablation_configs}
\footnotesize %
\begin{tabularx}{\textwidth}{@{}>{\raggedright\arraybackslash}X >{\raggedright\arraybackslash}X >{\raggedright\arraybackslash}X >{\centering\arraybackslash}X >{\raggedright\arraybackslash}X@{}}
\toprule
\textbf{Paradigm} & \textbf{Dataset} & \textbf{Budget level} & \textbf{Subjects (varied)} & \textbf{Trials per subject (varied)} \\
\midrule
\textbf{MI} & BNCI2014\_001 & High ($\sim$3k) & (6, 8) & (500, 375) \\
& & Low ($\sim$1.5k) & (3, 5) & (500, 300) \\
& Lee2019\_MI & High ($\sim$6k) & (30, 40, 50) & (200, 150, 120) \\
& & Low ($\sim$3k) & (15, 30, 50) & (200, 100, 60) \\
& Yang2025 & High ($\sim$10k) & (15, 20, 30) & (600, 450, 300) \\
& & Low ($\sim$5k) & (10, 15, 30) & (450, 300, 150) \\
\midrule
\textbf{P300} & BI2015a & High ($\sim$20k) & (15, 20, 30) & (1000, 750, 500) \\
& & Low ($\sim$10k) & (10, 20, 25) & (500, 250, 200) \\
& Huebner2017 & High ($\sim$75k) & (6, 10, 12) & (12500, 7500, 6250) \\
& & Low ($\sim$37.5k) & (3, 6, 10) & (12500, 6250, 3750) \\
& Huebner2018 & High ($\sim$84k) & (6, 8, 12) & (14000, 10500, 7000) \\
& & Low ($\sim$42k) & (3, 6, 10) & (14000, 7000, 4200) \\
\midrule
\textbf{SSVEP} & Kalunga2016 & High ($\sim$512) & (8, 12) & (64, 42) \\
& & Low ($\sim$256) & (4, 8) & (64, 32) \\
& Lee2019\_SSVEP & High ($\sim$4k) & (20, 25, 40) & (200, 160, 100) \\
& & Low ($\sim$2k) & (10, 20, 40) & (200, 100, 50) \\
& MAMEM2 & High ($\sim$600) & (6, 8) & (100, 75) \\
& & Low ($\sim$300) & (3, 6) & (100, 50) \\
\bottomrule
\end{tabularx}
\end{table}

\clearpage
\FloatBarrier

\clearpage
\FloatBarrier

\clearpage
\FloatBarrier

\clearpage
\FloatBarrier

\clearpage
\FloatBarrier

\clearpage
\FloatBarrier

\end{document}